\documentclass[journal]{new-aiaa}
\usepackage[utf8]{inputenc}

\usepackage{graphicx}
\usepackage{amsmath}
\usepackage[version=4]{mhchem}
\usepackage{siunitx}
\usepackage{longtable,tabularx}
\usepackage{hyperref}
\usepackage{siunitx}
\usepackage{graphicx}
\usepackage{tikz}
\usetikzlibrary{external}
\usetikzlibrary{positioning, fit, arrows.meta, shapes, decorations.pathreplacing}
\usetikzlibrary{spy}
\usepackage{pgfplots}
\pgfplotsset{compat=newest}
\pgfplotsset{plot coordinates/math parser=false}
\usepackage{xcolor}
\usepackage{pdfpages}
\usetikzlibrary{plotmarks}
\usepackage{todonotes}
\usepackage{caption}
\usepackage{subfigure}

\usepackage{changes}
\usepackage{cancel}

\title{Machine Learning based Prediction of \\Ditching Loads}

\footnotetext{Copyright © 2024 by the American Institute of Aeronautics and Astronautics, Inc. All rights reserved.}
\author{Henning Schwarz\footnote{Research Associate, Institute for Fluid Dynamics and Ship Theory, Am Schwarzenberg-Campus 4, henning.schwarz@tuhh.de (Corresponding Author)}, Micha Überrück \footnote{Research Associate, Institute for Fluid Dynamics and Ship Theory, Am Schwarzenberg-Campus 4, micha.ueberrueck@tuhh.de.}, Jens-Peter M. Zemke \footnote{Chief Engineer, Institute of Mathematics, Am Schwarzenberg-Campus 3, zemke@tuhh.de.} and Thomas Rung \footnote{Professor, Institute for Fluid Dynamics and Ship Theory, Am Schwarzenberg-Campus 4, thomas.rung@tuhh.de.} }
\affil{Hamburg University of Technology, D-21073 Hamburg, Germany}

\begin{document}
\newlength\figureheight
\newlength\figurewidth

\maketitle
\begin{abstract}
We present approaches to predict dynamic ditching loads on aircraft fuselages using machine learning. The employed learning procedure is structured into two parts, the reconstruction of the spatial loads using a convolutional autoencoder (CAE) and the transient evolution of these loads in a subsequent part. Different CAE strategies are assessed and combined with either long short-term memory (LSTM) networks or Koopman operator based methods to predict the transient behaviour.   
The training data is compiled by an extension of the momentum method of von Karman and Wagner and the rationale of the training approach is briefly summarised. 
The  application included refers to a full-scale fuselage of a DLR-D150 aircraft for a range of horizontal and vertical approach velocities at $\SI{6}{\degree}$ incidence. 
Results indicate a satisfactory level of predictive agreement for all four investigated surrogate models examined, with the combination of an LSTM and a deep decoder CAE showing the best performance.

\end{abstract}

\section{Introduction 
}
\label{sec:into}
\begin{figure}[h!]
        \centering
        \includegraphics[width=12cm]{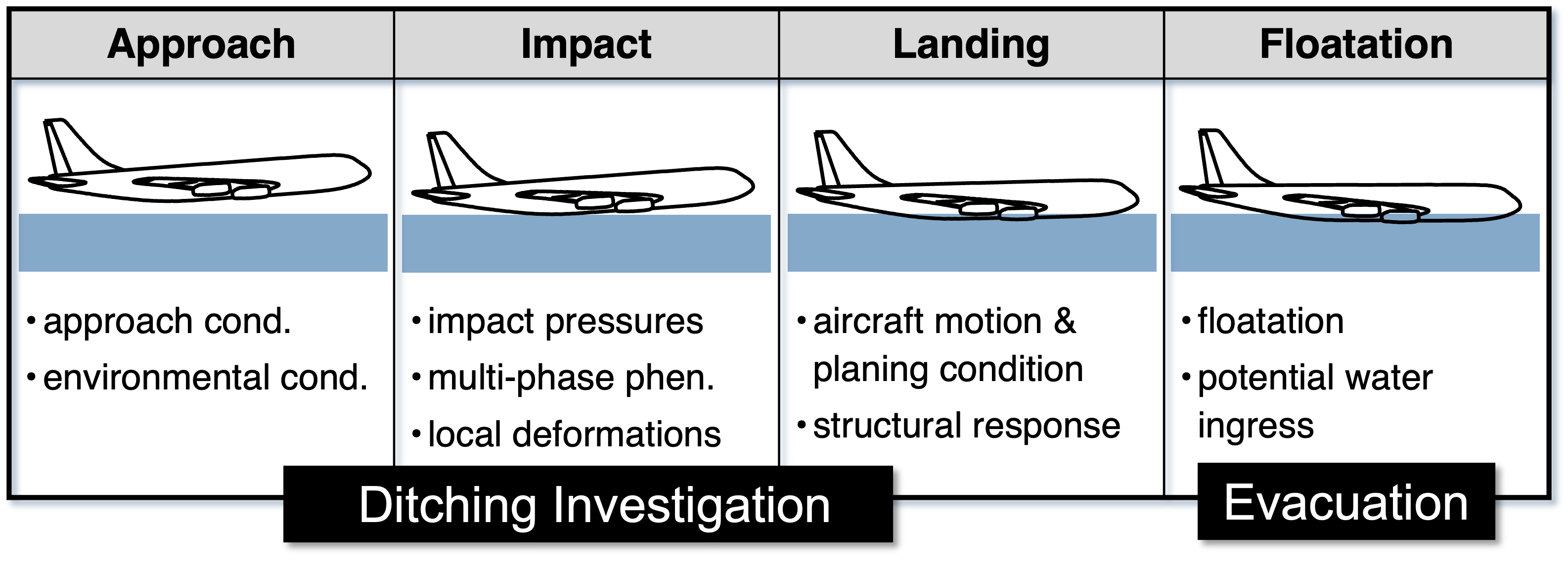}
        \caption{Frequently employed ditching investigation phases.}
        \label{fig:4phasen}
\end{figure}

 Ditching describes the emergency landing on water where several regulatory items have to be considered for large transport aircrafts. Resulting certification provisions are found in the EASA/FAA CS 25.563 ‘Structural ditching provisions’, CS 25.801 ‘Ditching provisions’ and CS 25.807(e) ‘Ditching emergency exit for passengers’. They aim at preventing immediate injuries for occupants and limiting aircraft damages to ensure the floatation time to be long enough for the occupants to exit 
 the aircraft safely. 
 Ditching is either investigated  by accident analysis, experiments, which are usually based on scaled model tests, or numerical simulations. 
 Investigations are typically split into four subsequent chronological phases, i.e., an approach, impact, landing and floating (evacuation) phase, cf. Fig. \ref{fig:4phasen}. The impact phase is the focus of the load  analysis, on which the initial conditions, such as the pitch of the aircraft, its forward and sink speeds, a possible fuel jettison or the sea state encounter angle, are the subject of optimization. 
 
 The majority of investigations follow a one-way fluid-structure coupling philosophy, and are partitioned into the determination of the hydrodynamic loads obtained for a (virtually) rigid body and a subsequent assessment of the structural response. 
 Model-scale ditching tests in combination with data from previously commissioned configurations are a  traditional way of load analysis. Conducting model scale experiments,  scaling laws and associated scaling effects have to be taken into account, and strong similarity of the load situation is hard to achieve. Because of the dominance of inertia and gravitational forces during the impact phase, experiments are usually performed using Froude-scaling for the model properties and impact conditions. Froude-scaling drastically reduces the test velocity and neglects the similarity of viscous forces. Moreover, the similarity of multi-phase  aspects, i.e., cavitation and ventilation, are not considered, e.g., \cite{zhang:2012}. Several authors have outlined the importance of multi-phase effects for a realistic load prediction and the corresponding trajectory of the aircraft, 
 particularly when severe longitudinal fuselage curvatures and high forward velocities are observed \cite{streckwall:2007}, and errors associated with this are difficult to correct. However, empirical corrections based on the results of a set of targeted experimental research efforts performed at CNR-INSEAN  \cite{iafrati:2021, iafrati:2019, spinosa:2021} 
 have recently become part of industrialized numerical tools. These  tools 
 aim at predicting the aircraft's motion and the global and local forces acting on the airframe during ditching.
The simulation method \emph{ditch} \cite{bensch:2003, Gropen14} is such a 
procedure and used for the present research. 
The method supports extensive sensitivity studies, optimizing the aircraft's approach conditions and serves a simulation-based design verification.
It is based on an extension of
the momentum method by von-Karman \cite{karman:1929} and Wagner \cite{wagner:1932},  in which  
the aircraft motion is restricted to 3 degrees of freedom (3DoF; horizontal, vertical and pitch motion) to reduce the computational effort.  
The method includes models for two-phase effects, sea state influences, aerodynamic loads, thrust and can take into account component losses and hydrodynamic niche loads. Due to the 3DoF limitation, only half of the aircraft is modeled, i.e., the (starboard) half fuselage, complemented by one wing with its engines and  horizontal tail part. 
 The predicted load histories of selected optimal or critical scenarios are 
 passed to a structural analysis for the verification of the structural integrity. 
The granularity and computational cost of the load prediction and the structural prediction differ by orders of magnitude, suggesting the use of one-way coupling. 

A medium term goal is to consider deformation-related load changes or even crash elements using a simplified two-way coupling.
One possible approach is to account for structural deformations using an ML-based surrogate model within a two-way coupling. 
The reconstruction of the loads shown in this article is a first step towards analyzing the potential of available ML strategies, and should 
scrutinize data-based methods for learning transient load evolutions obtained from the \emph{ditch} procedure. 
The considered ML strategy consists of performing load predictions for time series. The data dimension is initially reduced by a convolutional autoencoder (CAE). Subsequently, two different routes are pursued to advance the reduced
space 
in time. On the one hand, a long short-term memory (LSTM) \cite{hochreiter:1997} network is applied along the lines of recent fluid dynamics related publications \cite{eivazi:2020, wu:2021}, which follow strategies similar to the LED framework more recently proposed in \cite{vlachas:2022}. 
Alternatively, Koopman operator approaches 
\cite{takeishi:2017, lusch:2018, otto:2019} are utilized to predict the temporal load development, which may provide better access to interpretability than LSTM due to the linearity of the obtained operator. Strictly speaking, we however focus on a non-linear version that maps back to the full dimension after each prediction in the reduced space as this leads to more accurate results than the linear version in our case.
To motivate the use of these complex models, a simpler deterministic model is also considered, namely the dynamic mode decomposition (DMD) \cite{schmid:2010}. Before comparing the autoencoder-based models, we demonstrate that the ditching loads are not  predicted accurately and at the same time fast by DMD.
To the best of our knowledge, this is the first application of ML for the prediction of ditching loads and 
the aim to build 
surrogate models that approximate the structural response during ditching.

The remainder of the paper is structured as follows: in
Section \ref{sec-dmeth} the ditching load analysis method that provides the training data for this study is briefly described and illustrated using two test cases.
Section \ref{sec-mmeth} is dedicated to the utilized machine learning methods, in particular the respective approaches to model the transient load history.  
An application to a realistic DLR-D150 fuselage in full scale \cite{dlr136615} is discussed in Sections \ref{subsec:data} and \ref{sec:res} and final 
conclusions are drawn in the last section \ref{sec:concl}.

\section{Ditching Load Analysis Methodology}
\label{sec-dmeth}

The level of complexity of numerical ditching investigations ranges from (semi-) analytical water-impact formulas \cite{climent:2019} to geometrically resolving, 3D time-dependent methods taking into account the complete aircraft structure and the near field of the surrounding fluids \cite{siemann:2018}. To this effect, methods are commonly distinguished into high- and low-fidelity approaches. High-fidelity approaches are based upon general, high-resolution field methods for computational mechanics, featuring a mesh- or particle-based discretization of the aircraft and its surrounding domain, or hybrids of the latter in partitioned fluid-structure analysis \cite{streckwall:2007, siemann:2018}. They offer a broad predictive realm at the expense of a considerable computational cost. Fully resolved high-fidelity 3D simulations are usually only performed for final configurations covering a limited time frame of approximately one second. They are mostly restricted to the structural part of the analysis for the most critical rear fuselage, the loading of which is often derived from a prior rigid aircraft ditching analysis. 
Hereby computed pressure loads compiled for a final industrial configuration and relevant ditching scenarios are mapped onto the fuselage to perform a detailed structural analysis that typically involves several million degrees of freedom and several days of run-time on parallel machines.

Furthermore, high-fidelity load simulation methods are frequently used to train efficient low-fidelity simulation methods \cite{streckwall:2007}. The latter support investigating the complete ditching process during the design phase which involves optimizing various parameter influences on the load and the load history often during several hundred simulations. 

 It is evident that high-fidelity methods are not suitable to be used in the design phase to optimize the safety of new aircraft configurations or concepts. This motivates the development of low-fidelity approaches, which refer to combinations of low spatial resolution, simplified models or semi-analytical methods. Low-fidelity load simulation approaches usually only discretize the aircraft shape, often in a simplified manner, or refer to modified shapes that are accessible by analytical formulae \cite{lindenau:2009, delbuono:2021, Tassin13}, which facilitate run times of the order of seconds or minutes.
 Current interest exists to augment low-fidelity load simulation approaches with structural analysis elements
 to approximate fluid-structure interaction influences on the loads. 

\subsection{Ditching Simulation Method}
\begin{figure}[h!]
    \centering
    \includegraphics[width=15cm]{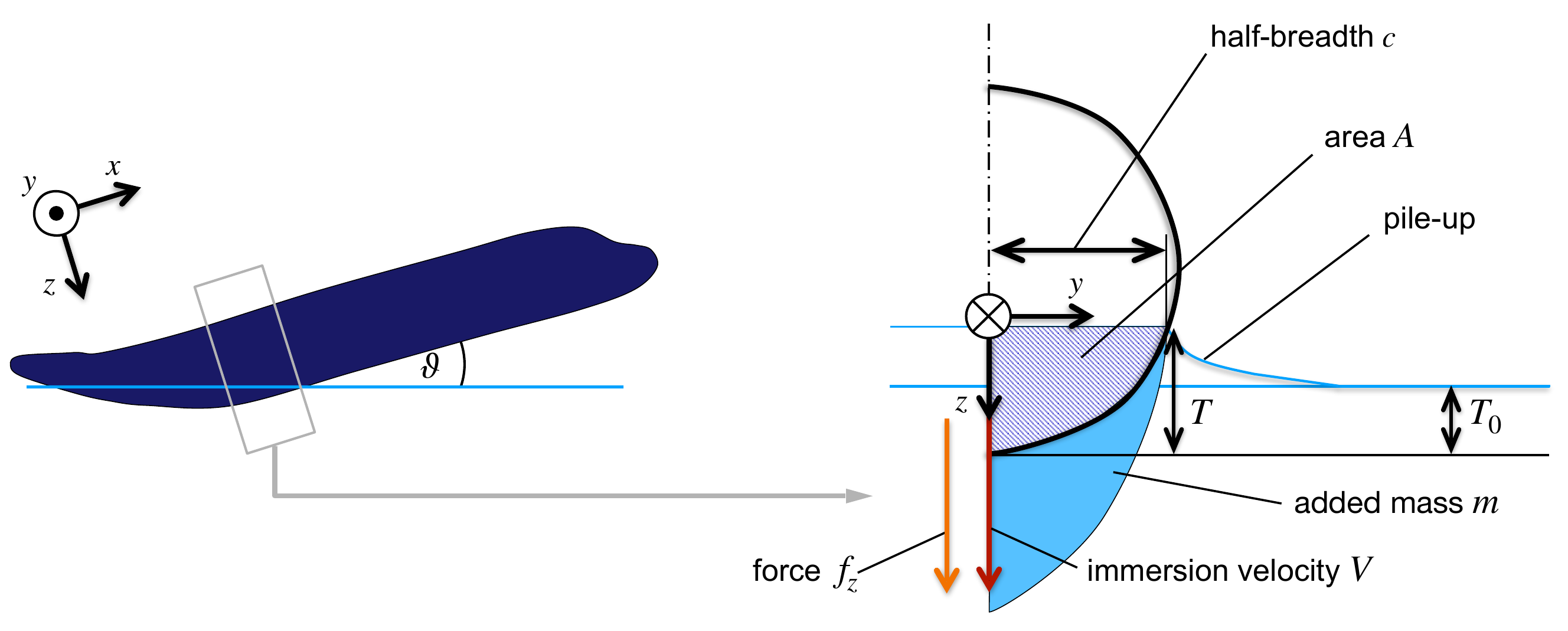}
    \caption{Aircraft cross section and the corresponding quantities for \emph{ditch} simulations.}
    \label{fig:crosssection}
\end{figure}

In this work the 2D+t method \emph{ditch} is used to provide the training data for hydrodynamic fuselage loads \cite{bensch:2003}. The method is essentially a strip method and is based on a momentum theory for vertical impact forces per unit length. The  aircraft geometry under consideration is first broken down into individual components, e.g., fuselage, wing, engine(s) and tail. Each component is subsequently subdivided into equidistant sections along the longitudinal direction $x$ of the aircraft,  which is inclined against the free surface by the pitch angle $\vartheta$,  and each section is then discretized along the section arc $s$, which is composed of the $y$- and $z$-coordinates, cf. Fig. \ref{fig:crosssection}. The approach assumes slender bodies with forward speed $u$ in the direction of $x$.
The dependent variables refer to the vertical (immersion) velocity $V$, the submerged half-breadth $c$, the submerged area $A$, the submergence $T$. These properties are assessed  directly at the respective section and are used to compute the vertical force per unit length $f_z$, i.e., the desired impact load.

The dynamic part of the vertical force per unit length and the change of the vertical velocity are related through conservation of vertical fluid momentum, i.e.,
$
 f_z \sim -  \, D\left( m_A \, V \right)/Dt 
$, where $m_A$ is the added fluid mass per unit length. For the example of half of a circular cylinder cross section, the added mass of a fluid with density $\rho$ refers to $m_A=(\rho \pi c^2)/2$, with $c$ being the half-breadth at the waterline including the pile-up during immersion. 
In a first step the submergence and submerged half-breadth and -area for every section are calculated according to Wagner \cite{wagner:1932} including an empirical approach for the water-line pile-up $T-T_0$ communicated by Söding in \cite{bensch:2003}. 
 For shallow submerged sections, an adequate approach to model the immersion velocity reads 
\begin{align}
    V = \frac{1}{c}\frac{DA}{Dt} - \frac{D}{Dt} \left(T-T_0\right),
\end{align}
where $T_0$ is the submergence of the undisturbed free surface and $T$ is the submergence including the pile-up.
The total vertical hydrodynamic force per unit length and the underlying pressure $p$ are computed from
\begin{align}
    f_z &= -k \rho \frac{\pi}{4} \left(c^2 \frac{DV}{Dt} + V\frac{Dc^2}{Dt}\right) - \rho g A_0 \label{eq:ditch-force} \\
    p &= k \rho \frac{DV}{Dt} \sqrt{c^2-y^{2}} + k \rho V c \frac{Dc}{Dt} \frac{1}{\sqrt{c^2-y^{2}}} + \rho g \zeta_0 \; . 
    \label{eq:ditch-press}
\end{align}
Here $g$ refers to the gravity constant and $k$ is a correction factor that accounts for blunt (non-circular) immersed cross section or deeper immersion, which is also used to close the waterline pile-up from $T-T_0= 0.6 k A/c$. Mind that the substantial derivative $D/Dt$ splits into a local ($\partial / \partial t$) and a convective part ($u\partial/\partial x$). 
The last terms of (\ref{eq:ditch-force}) and (\ref{eq:ditch-press}) refer to the hydrostatics in the case of an undisturbed water level, given by $A_0$ for the submerged half area and $\zeta_0$ for the submergence without pileup.

In addition, \emph{ditch} employs models for pressure/load modulations due to cavitation and ventilation, as well as more complex aspects (e.g., flow separation). These mostly limit the immersion velocity, the computed pressures or modify the correction factor $k$ depending on the dynamics of the aircraft. The hydrodynamic pressure loads are supplemented by viscous drag forces,  that relate to simple friction line characteristics, and local niche drag. Having integrated the hydrodynamic loads over the wetted aircraft sections, aerodynamic loads, external forces and engine thrust are optionally added to the resulting force vector $\vec{\mathbf{f}}$.
The employed aerodynamic model is quite simple. It is based upon a prescribed span-wise circulation distribution and employs the basic wing geometry. At the beginning of the approach phase, an equilibrium of weight and lift forces is assumed and  provides a lift coefficient.
  Subsequently, changes of the pitch angle (linearly) or the velocity yield to changes of the lift forces.

The force vector is applied to compute the acceleration $\ddot{\vec{\delta}}$ from Newtons second law of motion and to subsequently determine the motion and position of the aircraft, viz.
\begin{align}
    \left(\mathbf{M} + \mathbf{M}_A\right) \ddot{\vec{\delta}} = \vec{\mathbf{f}} \, ,
\end{align}
where $\mathbf{M}$ [$\mathbf{M}_A$] refer to the aircraft [added] mass and moment of inertia. 
An iterative solution process is employed in every time step to account for the dependence of the added mass $\mathbf{M}_A$ to the accelerations $\ddot{\vec{\delta}}$.

\subsection{Validation}
Since the method has been validated several times, we restrict the validation to two frequently used examples.

\subsubsection{Guided Impact of a Curved Plate}
The first validation case is concerned with  experimental data compiled during the EU-sponsored  \emph{SMart Aircraft in Emergency Situations (SMAES)} research project. 
Various experiments with flat and curved plates -- that represent an aircraft fuselage section -- were performed 
at a dedicated high-speed, guided-impact facility at the CNR-INM (INSEAN) in Rome. 
The general setup of all experiments, which vary in impact speed and pitch angle, consists of a track, where a trolley with the specimen equipped with sensors to measure acceleration, pressures and forces, accelerates  and 
 impacts into the water.

\begin{figure}[h!]
    \centering
        \includegraphics[width=\textwidth]{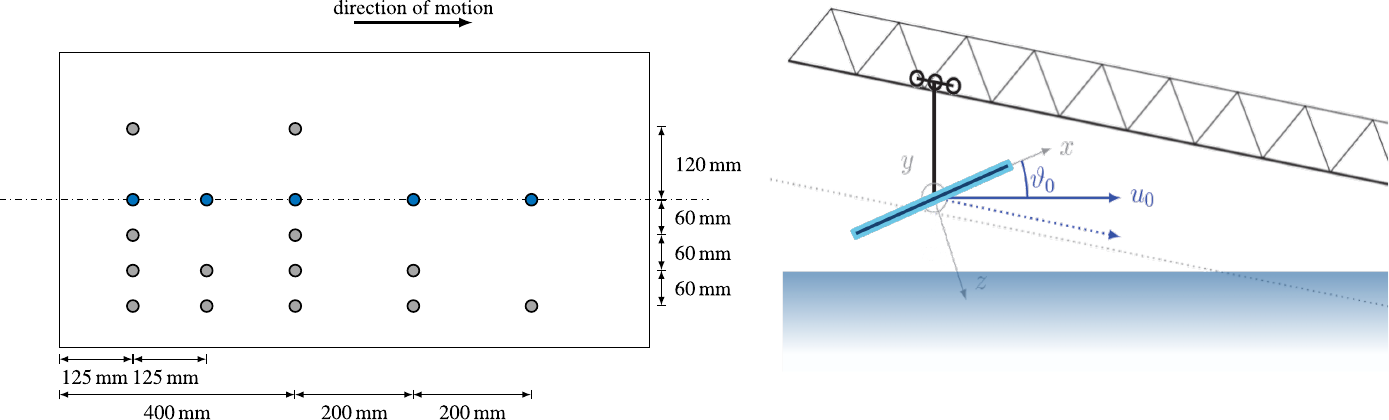}
    \caption{Illustration of the investigated guided impact of a curved plate case (left: outline of pressure sensor locations; right: sketch of the test rig).}
    \label{fig:smaes_1222}
\end{figure}

\begin{figure}
    \centering
    \includegraphics[width=12cm]{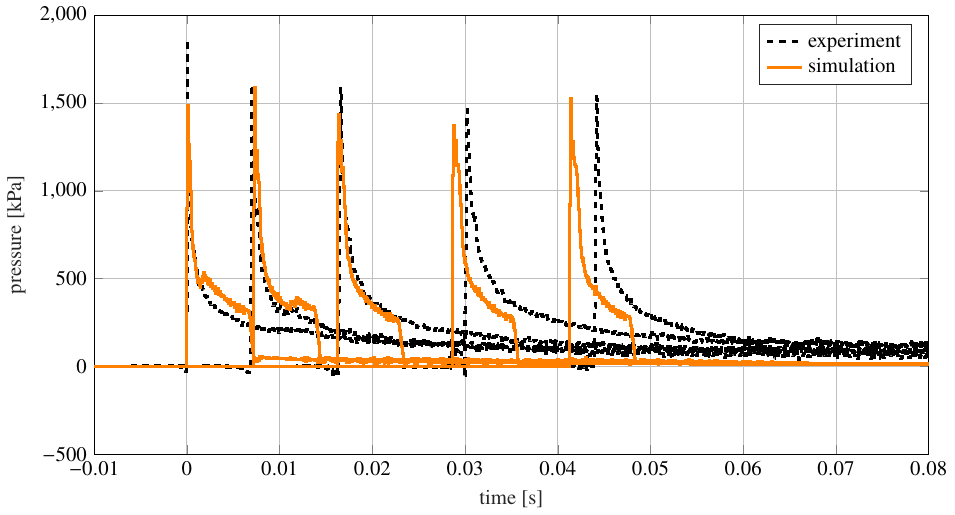}
    \caption{Comparison of predicted (\emph{ditch}) and measured pressures relative to the ambient pressure (sensor locations correspond to blue symbols in Fig. \ref{fig:smaes_1222}).}
    \label{fig:smaes_1222-r}
\end{figure}

The examined case is illustrated in Fig. \ref{fig:smaes_1222}. It refers to a curved plate, which has a length of $\SI{1.0}{m}$, a width $\SI{0.5}{m}$ and a transversal curvature radius of $\SI{2}{m}$.  The investigated impact velocities read $u_0= \SI{40}{m/s}$ in the horizontal  and $\SI{1.5}{m/s}$ in the vertical direction. The pitch angle refers to  $\vartheta_0 =\SI{6}{\degree}$ against the horizontal.
The spatial discretization of the computed geometry matches the location and the cross sectional area of pressure sensors and refers to $\SI{0.005}{m}$ along the circumferential (ds) and longitudinal direction (dx). 
The  time step reads $\Delta t = \SI{6e-5}{s}$, which returns an upper CFL number of approximately  $0.5$.
Emphasis is restricted to a comparison of predicted and measured impact pressure along the center line of the specimen. As outlined in Fig. \ref{fig:smaes_1222-r},  
the simulated peak loads and their times of occurrence agree with measured data for the first three sensors. The observed temporal deviations of the last two sensors can be attributed to the elastic motion of the track, induced by the high impact forces, cf. \cite{iafrati:2015}. 
The experimental [simulated] pressure signal drops to small values at approximately $\SI{0.06}{s}$ [$\SI{0.05}{s}$], when the water level reaches the upper boundary of the discretized geometry. Comparing simulations and measurements, minor differences in residual forces are observed, but do not significantly affect the global forces and motion. 

\subsubsection{Free Motion of a Model Fuselage}

\begin{figure}
    \centering
    \includegraphics[width=14cm]{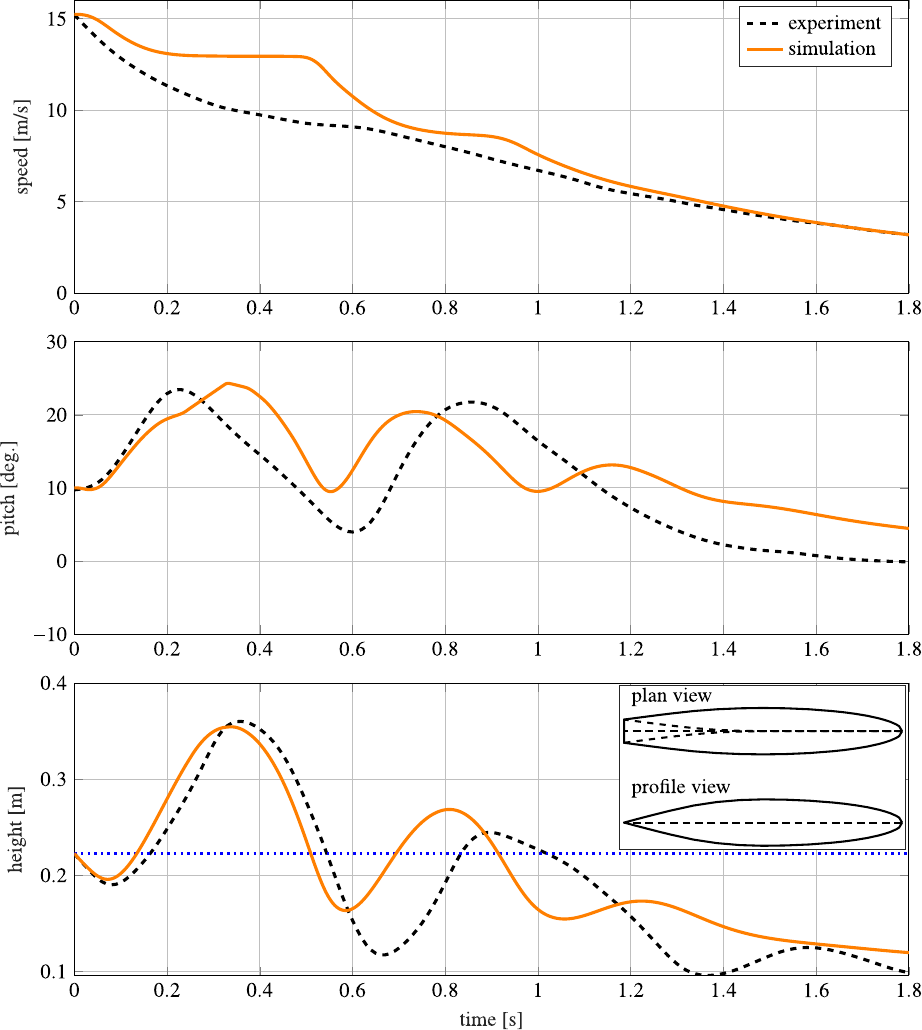}
    \caption{Comparison of predicted (\emph{ditch}) and measured motion of the NACA TN2929 fuselage model D. Side and top view of the fuselage model are depicted in the lower graph.}
    \label{fig:tn2929-d50}
\end{figure}

The frequently cited NACA Technical Note 2929 \cite{tn2929} is used as a second example. The focus of the experimental contribution was on investigations into the influence of the rear hull geometry on the ditching behaviour. The analyzed nine geometries range from model A to model J, and can be divided into three groups. Model A-C are basic bodies of revolution with a no  longitudinal curvature (A) and curvature due to sweeping down (B) or sweeping up (C) the rear. Model D-F refer to A-C but have a split center line, which widens the rear hull. Model G-J resemble stretched versions of A-C that feature a higher aspect ratio. 
The available measurements are limited to motion data analyzed from recorded images. These are afflicted by some uncertainty regarding aerodynamic influences and initial conditions, possible tear-off of the horizontal tail and inertia terms, roll and yaw motions that are not documented, and the sampling and filtering of the processed data.
We limit the data displayed here to a brief comparison of the predicted and measured planar motion for the model D shown in {the lower graph of Figure \ref{fig:tn2929-d50}} and one particular initial velocity.
The fuselage geometry is a cylindrical body with a split center line at the rear. It has a length of 
$L=\SI{1.219}{m}$ and a maximum radius of $R=\SI{0.1016}{m}$ corresponding to an aspect ratio of $L/(2R) = 6$. The fuselage is equipped with wings and horizontal tail and launched with an initial overall  
speed of $\SI{15.24}{m/s}$ at a pitch angle of $\SI{10}{\degree}$. The initial vertical velocity follows from the experimentally reported initial changes of the center of gravity (CoG) height above the free surface.

For comparison, the digitized data from the report and the predicted evolution of the horizontal speed, pitch angle and CoG height above the free surface are shown in Fig. \ref{fig:tn2929-d50}. 
The blue dotted line symbolizes the distance of the center of gravity from the free surfaces when the aircraft is in contact with water at 10$^\circ$ pitch. 
The displayed motion begins with the first impact on the water at $t=\SI{0}{s}$ 
and depends on the model's center of gravity position and the moment of inertia, which were extracted from the report \cite{tn2929}. 
As also described in the NACA report  \cite{tn2929}, the general behavior of the model shows a tendency to skip and the model leaves the surface twice before it enters the landing phase at $t\ge 0.9$ in the simulation, which would not be desirable for real aircrafts.

A few things of the comparison draw attention. The gradient of the horizontal speed is very much different in the initial situation though the pitch is fairly similar in this phase.
Moreover, the simulation displays two plateaus, which are much less pronounced in the experiment. Both correspond to the time, where the model leaves the surface
 and the decelerating  hydrodynamic drag force is missing in the simulations. 
 The experimental data largely confirm the corresponding motion of the body, see pitch and height motion graphs, but at the same time indicate aerodynamic braking forces that the simulation model does not inhere.
Beyond this, a phase shift between experiment and simulation is observed for the height and pitch motion. Both are closely connected and are explained with uncertainties of the mass distribution and the strength as well as the point of application of aerodynamic forces.
Nevertheless, the ditching behavior is well reproduced. Both skipping phases are captured, the maximum pitch angle is nearly the same and the deceleration in the overall time is similar.

\section{Machine Learning Methods}
\label{sec-mmeth}
The surrogate models used herein aim at predicting the temporal evolution of the spatial loads on the fuselage of an aircraft during the impact phase of a ditching event, although the ultimate goal of our research is to approximate structural deformations.
The two considered spatial coordinates refer to the longitudinal coordinate ($x$) measured from the tail of the aircraft and the circumferential coordinate ($s$). The spatial data is extracted from a regular grid, which typically involves around 20 000 nodes, cf. Sec. \ref{subsec:data}.
The spatial dimensionality is initially reduced using a CAE, preserving spatial dependencies, to extract low dimensional features of the full order data and 
reduce the input dimension for the layers focusing on the temporal load evolution. This serves to reduce the necessary amount of parameters for those layers and, as a result, accelerate the training procedure.
The temporal load evolution based on the reduced data is described by either an LSTM network or an approximation to the Koopman operator~\cite{Koopman:1931}, and differences in these respects are part of this works scope.
Though the approach formally involves two  subsequent steps, i.e., 
 the dimension reduction and the temporal evolution, present results exclusively refer to a joint training of both components. This turns out to be superior to a separate training of the components, see also~\cite{wu:2021,lusch:2018, otto:2019}, as the sequential minimization of the spatial reduction loss and the temporal prediction loss usually 
 exceeds a combined minimization. 
Due to the joint construction, the CAE will however not be restricted to the dimension reduction and could also support the temporal evolution, cf. Subsection \ref{sec:ae_lstm}. All presented models are built in TensorFlow~\cite{TF:2016} using the Keras framework \cite{chollet2015keras}.

 \smallskip
Autoencoder and LSTM networks have recently been used in a different setting by Lazzara et al. \cite{lazzara:2022} to design a surrogate model for predicting the landing loads of an aircraft. 
Their approach differs in the use of 1D+t data, i.e., only global loads and accelerations were computed in \cite{lazzara:2022},  
compared to our 2D+t data as well as the overall strategy of their networks. An LSTM-autoencoder that comprises multiple LSTM layers with a decreasing and increasing dimension of the hidden states in the encoder and decoder, respectively, is utilized to reconstruct the complete time series in one shot in \cite{lazzara:2022}. An additional feed forward neural network maps the 26 input parameters of the high-fidelity model, i.e., pitch angle, dynamic pressure, landing gear shock absorber characteristics, aerodynamic center, center of gravity, moments of inertia, and three parameters defining the mass distribution, to the corresponding latent space representation in line with previous works, e.g., \cite{Agostini20, Swischuk.2019, Pache22}. This enables a reconstruction of the complete 1D time series at once from a small set of input parameters.
Aiming at a two-way coupling, our networks intend to sequentially predict the local data by a time advancing surrogate model simulation.

\subsection{Convolutional Autoencoder (CAE)}
\begin{figure}[!h]
     \centering
     \includegraphics{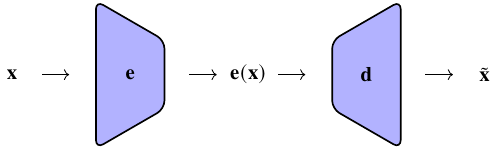}
    \caption{Scheme of an autoencoder.}
    \label{fig:ae}
\end{figure}

 An autoencoder is a neural network that reconstructs the input data after passing it through a lower dimensional \emph{latent space}.
 The reduction of the data dimension is performed by applying an encoder
 $\mathbf{e}: \mathbb{R}^{n} \rightarrow \mathbb{R}^{m} $ to map the $n$-dimensional input data on to the $m$-dimensional latent space, with $m\ll n$. The decoder $\mathbf{d}: \mathbb{R}^{m} \rightarrow \mathbb{R}^{n}$ subsequently maps the data back to the original space, often using a symmetric structure to the encoder, cf. Fig. \ref{fig:ae}. During training, the input and the output are identical. Thus, with $\tilde{\mathbf{x}}=\mathbf{d}(\mathbf{e}(\mathbf{x}))$ the input $\mathbf{x}$ is first encoded and afterwards the encoded data is decoded while demanding that the output is as close to the original data as possible. This way, an autoencoder extracts the most important features of the input data inside the latent space and can be used for dimension reduction.

Spatial dependencies, which are often crucial in data that describes physical results, e.g., solutions of partial differential equations,  can be captured by using convolutional layers. 
As the data passes through these layers, convolutions or, more specifically, cross-correlations are applied to them with custom-sized filters.
In a convolutional autoencoder (CAE), the encoder uses strided convolutions and/or pooling layers to successively decrease the dimension of the input.  
Pooling aims at extracting characteristic features of the spatial domain, e.g., extreme values of small regions, and the stride describes the step size of the filter, which is used to reduce the dimension.  
The decoder in a CAE uses strided transposed convolutions and/or upsampling to increase the dimension to the original size.
As the considered load data in this work 
involves spatial dependencies along the longitudinal and circumferential fuselage coordinates, a CAE will be the basis of all our models. 

\subsubsection{Non-local blocks}
Using convolutional layers to capture spatial dependencies, it might be difficult to consider global dependencies since the convolution kernel only captures local information. To solve this issue, Wang et al. \cite{wang:2017} proposed so-called \emph{non-local blocks}, which are related to \emph{self-attention} \cite{vaswani:2017}. 
The currently modeled ditching load data displays large spatio/temporal regions where the dynamic pressure load globally vanishes, e.g., at  later times and in non-wetted regions. Inspired by \cite{wu:2021}, we assessed the non-local blocks influence on our application. To this end, non-local blocks are applied subsequently to the last two of the convolutional layers in the encoder, cf. Subsection \ref{sec:ae_lstm} and Table \ref{tab:jointmodel1}. Using channels last and without including the minibatch size, the output of a 2D convolutional layer is of shape $H  \times W  \times C$,  where the different dimensions denote the height, the width and the number of channels of the output, respectively.
Non-local blocks make use of the \textsf{softmax} function $\textsf{softmax}: \mathbb{R}^n \rightarrow (0,1)^n$, viz. 
\begin{align}
    \textsf{softmax}(\mathbf{x})_i = \frac{e^{x_i}}{\sum_{j=1}^n e^{x_j}}. 
\end{align}
To reduce the efforts, linear embeddings $g$, $\theta$ and $\phi$ of the input $\mathbf{x}\in \mathbb{R}^{H  \times W  \times C}$ are computed before the \textsf{softmax} function is applied.
The embeddings are implemented as 2D convolutional layers without bias with $C/2$ filters of size $1\times 1$ (introduced in \cite{lin2014networknetwork}). They confine the number of output channels to $C/2$ which also reduces computational effort for the following operations \cite{wang:2017, wu:2021}.
The outputs of $g$, $\theta$ and $\phi$ are flattened along the spatial dimensions $H$ and $W$, such that we have $g(\mathbf{x})\in\mathbb{R}^{H\cdot W \times C/2}$, $\theta(\mathbf{x})\in\mathbb{R}^{H\cdot W \times C/2}$ and $\phi(\mathbf{x})\in\mathbb{R}^{H\cdot W\times C/2}$.
The output $\mathbf{z} \in \mathbb{R}^{H  \times W  \times C}$ of a non-local block is computed by 
\begin{align}
    \mathbf{y} &= \textsf{softmax}\left(\theta(\mathbf{x})\phi(\mathbf{x})^T\right)g(\mathbf{x}), &
    \mathbf{z} &= w(\mathbf{y}) + \mathbf{x}. 
    \label{eq:nonlocalblock}
\end{align}
By considering all entries of $\mathbf{x}$, global dependencies are considered in the calculation of $\mathbf{y}\in\mathbb{R}^{H\cdot W \times C/2}$ \cite{wang:2017}. To enable the residual form in Eq. \eqref{eq:nonlocalblock}, the number of channels of $\mathbf{y}$ is again increased to $C$ using $w$, which is a convolutional layer without bias and with $C$ filters of size $1\times 1$. Before $w$ is applied, $\mathbf{y}$ is reshaped to $H \times W \times C/2$.

\subsection{Long Short-Term Memory (LSTM)}
\begin{figure}[!h]
     \centering
	\includegraphics{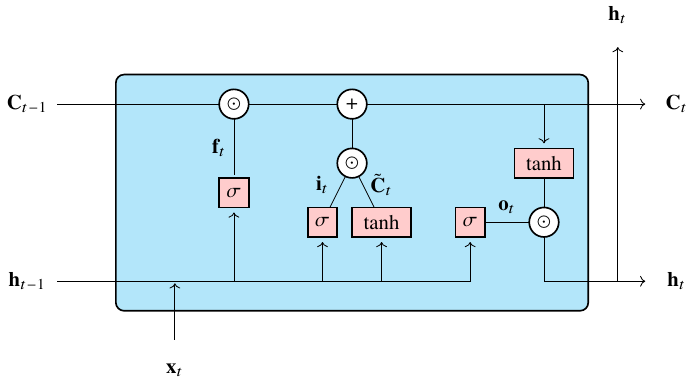}
    \caption{Structure of an LSTM cell.} 
    \label{fig:lstm}
\end{figure}

Long short-term memory (LSTM) networks \cite{hochreiter:1997} are recurrent neural networks (RNN). RNN are 
able to process sequences of data of varying lengths, i.e., suitable for time series prediction. Simple RNN output at time step $t$ a hidden state $\mathbf{h}_t\in\mathbb{R}^n$ that is used as additional input in the next step. LSTM have been designed to learn long-term dependencies with the aid of an additional cell state $\mathbf{C}_t\in\mathbb{R}^n$ of the same shape as $\mathbf{h}_t$. The output of an LSTM equals the hidden state. 
 Fig.~\ref{fig:lstm} depicts the structure of an LSTM cell which 
 contains five layers, colored in red. We employ the cuDNN library on GPU hardware, which recommends to use  
  the logistic function $\sigma:\mathbb{R}\to(0,1)$, $\sigma(x)=1/(1+e^{-x})$ and the hyperbolic tangent $\tanh:\mathbb{R}\to(-1,1)$ for performance reasons.  
  The three layers based on the logistic function are known as gates, these are the \emph{forget gate}, the \emph{input gate}, and the \emph{output gate} with output $\mathbf{f}_t\in\mathbb{R}^n$, $\mathbf{i}_t\in\mathbb{R}^n$, and $\mathbf{o}_t\in\mathbb{R}^n$, respectively,
\begin{align}
  \label{eq:LSTM_gates}
  \mathbf{f}_t
  &= \sigma\left(\mathbf{W}_{f,x}\mathbf{x}_t+\mathbf{W}_{f,h}\mathbf{h}_{t-1}+\mathbf{b}_f\right),
  & \mathbf{i}_t
  &= \sigma\left(\mathbf{W}_{i,x}\mathbf{x}_t+\mathbf{W}_{i,h}\mathbf{h}_{t-1}+\mathbf{b}_i\right),
  &\mathbf{o}_t
  &= \sigma\left(\mathbf{W}_{o,x}\mathbf{x}_t+\mathbf{W}_{o,h}\mathbf{h}_{t-1}+\mathbf{b}_o\right).
\end{align}
The entries in these vectors lie in $(0,1)$ due to the logistic function. Similarly, a candidate cell state $\widetilde{\mathbf{C}}_t$ is computed and used for the update of the cell state and the hidden state,
\begin{align}
  \label{eq:LSTM_update}
  \widetilde{\mathbf{C}}_t
  &= \tanh\left(\mathbf{W}_{c,x}\mathbf{x}_t+\mathbf{W}_{c,h}\mathbf{h}_{t-1}+\mathbf{b}_c\right),
  &\mathbf{C}_t &= \mathbf{C}_{t-1}\odot\mathbf{f}_t + \widetilde{\mathbf{C}}_t\odot\mathbf{i}_t,
  &\mathbf{h}_t &= \tanh(\mathbf{C}_t)\odot \mathbf{o}_t,
\end{align}
where $\odot$ denotes component-wise multiplication. An entry close to zero in $\mathbf{f}_t$ is used to “forget” information in $\mathbf{C}_{t-1}$, an entry close to one in $\mathbf{i}_t$ results in the addition of the corresponding information in the candidate cell state. The components of the new cell state, activated by the hyperbolic tangent, are selected for output based on $\mathbf{o}_t$. In TensorFlow/Keras, the weight matrices are grouped into an input weight matrix $\mathbf{W}_x$ and a recurrent weight matrix $\mathbf{W}_h$,
\begin{align}
  \label{eq:LSTM_grouping}
  \mathbf{W}_x &=
  \begin{pmatrix}
    \mathbf{W}_{i,x} \\ \mathbf{W}_{f,x} \\ \mathbf{W}_{c,x} \\ \mathbf{W}_{o,x}
  \end{pmatrix}\in\mathbb{R}^{4n\times m}, &
  \mathbf{W}_h &=
  \begin{pmatrix}
    \mathbf{W}_{i,h} \\ \mathbf{W}_{f,h} \\ \mathbf{W}_{c,h} \\ \mathbf{W}_{o,h}
  \end{pmatrix}\in\mathbb{R}^{4n\times n}, &
  \mathbf{b} &=
  \begin{pmatrix}
    \mathbf{b}_{i} \\ \mathbf{b}_{f} \\ \mathbf{b}_{c} \\ \mathbf{b}_{o}
  \end{pmatrix}\in\mathbb{R}^{4n},
\end{align}
where $m$ is the number of input features, $\mathbf{x}_t\in\mathbb{R}^m$. The input weight matrix is initialized using Glorot uniform initialization~\cite{glorot:2010}, the recurrent matrix is initialized with random orthonormal columns~\cite{SaxeMcClellandGanguli:2013}. The bias is initialized by $0$ with the exception of the forget gate initialized by $1$, since the cell state does not contain anything useful at the start of the training. We remark that the implementation of \eqref{eq:LSTM_gates}, \eqref{eq:LSTM_update}, and \eqref{eq:LSTM_grouping} is transposed in TensorFlow/Keras.

\subsection{Autoencoder-LSTM Models}
\label{sec:ae_lstm}
Inspired by Eivazi et al. \cite{eivazi:2020} and Wu et al. \cite{wu:2021}, 
different variants of an autoencoder-LSTM combination are assessed in the present work.

Eivazi et al. \cite{eivazi:2020} did use a pipeline model, where both networks  
(autoencoder, LSTM) are trained separately. Following the procedure outlined in  \cite{eivazi:2020}, the autoencoder should initially be trained  with the training data at hand. Subsequently, the encoder part $\mathbf{e}$ of the autoencoder is employed to train the LSTM network. To this end, sequences of $\ell$ data snapshots $\mathbf{x}_{t-\ell+1}, \dots, \mathbf{x}_t$, which represent $\ell$ consecutive time steps, are encoded and their latent representations are used as input for the LSTM network, whose output is the following time step  $(t+1)$. As regards the output dimension of the LSTM network two options are conceivable: (I) to predict the reconstructed full order data $\mathbf{x}_{t+1}$ or (II) to predict the reduced latent representation $\mathbf{e}(\mathbf{x}_{t+1})$, the latter subsequently requires 
the decoder part $\mathbf{d}$ of the autoencoder 
to map the latent space prediction back to the full dimension. In the work of Eivazi et al. \cite{eivazi:2020}, the first variant (I) displays superior performance in conjunction with a pipeline model. 
Wu et al. \cite{wu:2021} propose to use a CAE with an LSTM network in a joint model to reduce propagating errors from the single components. In this case, the components of both the CAE and the LSTM network are part of a single neural network to be trained. After the LSTM network, a convolutional decoder follows to map the latent representation of the subsequent time step back to the full dimension. Furthermore, including multiple non-local blocks to capture global dependencies in the data improves the performance of their model. 

In this paper, 
we investigate three different convolutional joint models. More specific, we compare the performance of variant I and variant II in a joint model and also scrutinize the influence of including non-local blocks. 
In line with Wu et al. \cite{wu:2021}, we use strided (transposed) convolutions to decrease [increase]  the dimension in the encoder  [decoder]. 
The first convolutional joint model (CJM) follows variant I and the structure is depicted in Table \ref{tab:jointmodel1}.
\begin{table}[h!]
\caption{CJM[CJMNLB] structure. Conv2D($f$, $k$, $s$) denotes a 2D convolutional layer with $f$ filters of kernel size $k\times k$ and stride $s$ in both dimensions. The shapes do not include the minibatch size. The network is trained on $3$ consecutive grayscale patches of shape $128\times128$, we use channels last, thus the input is of shape $3\times128\times128\times1$. 
We keep the three time slices separate before the LSTM layers. The first LSTM layer returns the hidden states of all time steps, the second LSTM layer returns the hidden state of the last time step. Mind that the non-local blocks are used only by the CJMNLB and not the CJM.}
\centering
\begin{tabular}{cccc}
    \hline
    \hline
    Layer & Output shape & Layer & Output shape\\
    \hline
     Input & (3,128,128,1) &  Flatten() & (3,4096) \\
     Conv2D(8, 3, 2) & (3,64,64,8) & Dense(10) & (3,10) \\
     Conv2D(16, 3, 2) & (3,32,32,16) & LSTM(100) & (3,100) \\
     Conv2D(32, 3, 2) & (3,16,16,32) & LSTM(100) & (100) \\
     \quad \big[Non-local block\big] & \big[(3,16,16,32)\big] & Dense(16384) & (16384) \\
     Conv2D(64, 3, 2) & (3,8,8,64) & Reshape(128,128) & (128,128) \\
     \quad \big[Non-local block\big] & \big[(3,8,8,64)\big] \\
     \hline
     \hline
\end{tabular}
\label{tab:jointmodel1}
\end{table}
The impact of LSTM variant II in a joint model is investigated combining the convolutional joint model with a deep decoder (CJMDD) shown in Table \ref{tab:jointmodel2}.
\begin{table}[h!]
\caption{CJMDD structure. Conv2D($f$, $k$, $s$) and Conv2DT($f$, $k$, $s$) denote a 2D convolutional layer and 2D transposed convolutional layer, respectively, with $f$ filters of kernel size $k\times k$ and stride $s$ in both dimensions. The shapes do not include the minibatch size. The network is trained on $3$ consecutive grayscale patches of shape $128\times128$, we use channels last, thus the input is of shape $3\times128\times128\times1$. 
We keep the three time slices separate before the LSTM layers. The first LSTM layer returns the hidden states of all time steps, the second LSTM layer returns the hidden state of the last time step.}
\centering
\begin{tabular}{cccc}
    \hline
    \hline
    Layer & Output shape & Layer & Output shape\\
    \hline
 Input & (3,128,128,1) & Dense(10) & (10)\\
     Conv2D(8,3,2) & (3,64,64,8) & Dense(4096) & (4096)\\
     Conv2D(16,3,2) & (3,32,32,16) & Reshape(8,8,64) & (8,8,64)\\
     Conv2D(32,3,2) & (3,16,16,32) & Conv2DT(32,3,2) & (16,16,32) \\
     Conv2D(64,3,2) & (3,8,8,64) & Conv2DT(16,3,2) & (32,32,16) \\
     Flatten() & (3,4096)  & Conv2DT(8,3,2) & (64,64,8) \\  
     Dense(10) & (3,10) & Conv2DT(1,3,2) & (128,128,1) \\
    LSTM(100) & (3,100) & Reshape(128,128) & (128,128) \\
    LSTM(100) & (100)  &  \\
    \hline
    \hline
\end{tabular}
\label{tab:jointmodel2}
\end{table}
Finally, the convolutional joint model with non-local blocks (CJMNLB) is studied to assess non-local block influences, which includes non-local blocks into the CJM. As indicated in Table \ref{tab:jointmodel1}, the first non-local block is used after the third convolutional layer. The second non-local block follows the fourth convolutional layer. Both non-local blocks use half as many filters as the prior convolutional layer, 16 and 32 respectively, for the convolutions inside the non-local block. This bisection of the filter size inside the non-local block is also depicted in \cite[Figure 2]{wang:2017}. Like in the work of Wu et al. \cite{wu:2021}, 2D convolutional layers are applied inside the non-local block.
We note that non-local blocks could also be included in the encoder and decoder of the CJMDD. However, to limit the discussion, in this work we only use non-local blocks as an example in combination with the CJM.

The structures and parameters of the models are tuned by hand. In all models, $\ell=3$ input time steps are used to predict the next one.
This choice was made as it lead to better results compared to $\ell=1$ and $\ell=2$ in our experiments,  while not requiring too many input time steps to start the models.
Table \ref{tab:time_delay_model_error_comparison} shows the behavior of the overall average error of the CJMDD with increasing the number of input time steps, obtained from a sequence of 30 ditching test cases. The tabulated error measure and the test cases used are explained in more detail in Section  \ref{sec:res}.
Using more input time steps improves the performances further, but also increases the initial effort since more time steps have to be generated to start the models.
\begin{table}[h!]
\caption{Behavior of the total average error (cf. Subsection \ref{subsec:load_prediction}) obtained by the CJMDD for 30 test cases when varying the number of input time steps.}
\centering
\begin{tabular}{cc}
    \hline
    \hline
    Number of input time steps & Total average error\\
    \hline
    1 & 0.027\\
    2 & 0.022\\
    3 & 0.020\\
    4 & 0.019 \\
    5 & 0.016\\
    10 & 0.010\\
    \hline
    \hline
\end{tabular}
\label{tab:time_delay_model_error_comparison}
\end{table}
The three time slices, which are each reduced to a latent representation of 10 neurons, are kept separate before entering the LSTM network.
The latent space dimension of 10 was chosen as it did provide satisfactory results in our experiments while substantially compressing the data. Increasing the latent space dimension towards, for example, 50 for the CJMDD, leads to an error reduction of approximately 10\%. However, a small latent space might be advantageous to gain interpretability of the applied models in future work.
The LSTM network consists of two LSTM layers and a following dense layer. From the first LSTM layer, the hidden states of all time steps act as input for the second LSTM layer. The latter returns only the last hidden state. We initialize the parameters of the LSTM layers as mentioned in the prior subsection. 
The weights and biases of the other layers are also initialized using the default initialization of the network parameters in TensorFlow/Keras, which is the 
Glorot uniform initialization 
for the weights and zero initialization for the biases. The layers in the encoder and the decoder are activated using \textsf{LeakyReLU} \cite{maas:2013} with slope parameter $\alpha=0.01$. 
The dense layer following the LSTM layers as well as the last transposed convolutional layer of the CJMDD are linear layers and no activation function is applied.
The last dense layer of the CJM and CJMNLB being linear could seem to 
induce higher errors due to the low rank of the weight matrix. We note that this is not the case, as outlined in Subsection \ref{subsec:load_prediction}.

For the training of the networks, the mean squared error (MSE) from Eq. \eqref{eq:mse} is minimized using the Adam optimizer \cite{Kingma:2015} with a minibatch size of 128.
\begin{align}
        \mathrm{MSE}(\mathbf{y},\hat{\mathbf{y}}) := \frac{1}{n}\lVert \mathbf{y}-\hat{\mathbf{y}} \rVert^2_2 =  \frac{1}{n}\sum_{i=1}^{n}{(y_i-\hat{y}_i)^2}, \quad \mathbf{y}\in \mathbb{R}^{n}, \hat{\mathbf{y}} \in \mathbb{R}^{n}.
        \label{eq:mse}
\end{align}
The derivatives needed in the optimizer are computed with the backpropagation \cite{Rumelhart:1986} algorithm. 
By using joint models we drop the restriction that the CAE and LSTM are solely responsible for the dimension reduction and the temporal evolution, respectively. This loss function therefore acts on the output of the full models like in \cite{wu:2021}.

\subsection{Koopman Autoencoder (KAE)}

In the last decades, Koopman theory~\cite{Koopman:1931} has gained attention in data-driven modeling. For a function $\mathbf{f}: D \rightarrow D$ with $D \subset \mathbb{R}^n$,
a dynamical system of the form  
 \begin{align}
     \mathbf{x}_{t+1} = \mathbf{f}(\mathbf{x}_t), \quad \mathbf{x}_0 \in \mathbb{R}^n,
 \end{align}
 and a function space $F$ containing observable functions $\mathbf{g}: D \rightarrow \mathbb{C}$, the infinite dimensional Koopman operator $\mathcal{K}$ is defined as the composition of an observable function $\mathbf{g}\in F$ and \textbf{f}: 
 \begin{align}
     \mathcal{K}(\mathbf{g}) = \mathbf{g}\circ \mathbf{f}
    .
 \end{align}
It can easily be verified that $\mathcal{K}$ is linear. Using the Koopman operator, it is therefore possible to model a nonlinear dynamical system with a linear operator when the state variables are transformed to an infinite dimensional function space. The linear operator $\mathcal{K}$ can then be analyzed in order to analyze the dynamics of the system.

In practice, the Koopman operator can only be approximated in a finite dimensional space. 
Well-known ways to approximate the Koopman operator include the dynamic mode decomposition (DMD) \cite{schmid:2010}, where \textbf{g} is assumed to be the identity, and the extended dynamic mode decomposition (eDMD) \cite{williams:2015}. The latter uses a dictionary of nonlinear functions to augment the training data with the goal to capture nonlinearities, before basically performing DMD.
However, a good choice of the dictionary functions is important to get a good approximation of $\mathcal{K}$ and to prevent overfitting \cite{otto:2019}. 

\begin{figure}[!h]
     \centering
	\includegraphics{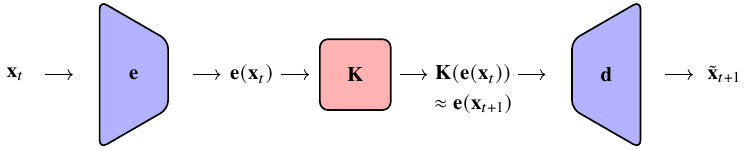}
    \caption{Scheme of a KAE.}
    \label{fig:kae}
\end{figure}

Recently, different strategies to apply autoencoders to approximate $\mathcal{K}$ have been presented \cite{takeishi:2017,lusch:2018, otto:2019,azencot:2020}. In these so-called \emph{Koopman autoencoders} (KAE), the main idea is to approximate $\mathcal{K}$ in the latent space to predict future time steps there. A vanilla architecture, comprising an encoder \textbf{e}, a linearity layer \textbf{K} for predicting the subsequent time step, and a decoder \textbf{d}, is illustrated in Fig. \ref{fig:kae}.  
The loss function consists of a reconstruction loss, a prediction loss as well as a linearity loss. For the current and the future time steps $\mathbf{x}_t$ and $\mathbf{x}_{t+1}$, it reads:  
\begin{multline}
    \mathrm{Loss_{VanillaKAE}}(\mathbf{x}_t; \mathbf{x}_{t+1}) = \\ \alpha_{\textsf{reconst}}\cdot \underbrace{\mathrm{MSE}\left(\mathbf{x}_t, \mathbf{d}(\mathbf{e}(\mathbf{x}_t))\right)}_{\text{Reconstruction loss}}
    + \alpha_{\textsf{predict}} \cdot \underbrace{\mathrm{MSE}\left(\mathbf{x}_{t+1}, \mathbf{d}(\mathbf{K}(\mathbf{e}(\mathbf{x}_t)))\right)}_{\text{Prediction loss}}
    + \alpha_{\textsf{linear}}\cdot \underbrace{\mathrm{MSE}\left(\mathbf{e}(\mathbf{x}_{t+1}), \mathbf{K}(\mathbf{e}(\mathbf{x}_t))\right)}_{\text{Linearity loss}}, 
    \label{eq:kaeloss}
\end{multline}
with non-negative hyperparameters $\alpha_{\textsf{reconst}}$, $\alpha_{\textsf{predict}}$, and $\alpha_{\textsf{linear}}$. The reconstruction loss term ensures that encoder and decoder form an autoencoder. The prediction loss term enforces that the network components jointly predict a future time step. The linearity loss term enforces linear dynamics in the latent space via $\mathbf{K}(\mathbf{e}(\mathbf{x}_t))\approx \mathbf{e}(\mathbf{x}_{t+1})$. 
In this model, the transformation \textbf{g} is learned by the encoder and the decoder.  
Multiple options exist to further enhance the model. 
For example, Lusch et al. \cite{lusch:2018} include regularization terms in their loss function. By demanding $\mathbf{K}^m(\mathbf{e}(\mathbf{x}_t))\approx \mathbf{e}(\mathbf{x}_{t+m})$ as well as $\mathbf{d}(\mathbf{K}^m(\mathbf{e}(\mathbf{x}_t)))\approx \mathbf{x}_{t+m}$, the linearity and the predictive behaviour are forced over multiple time steps. This means that one time step can be used to predict multiple future time steps. Furthermore, they propose to use an auxiliary network to parameterize the eigenvalues of \textbf{K}, which is forced to have block diagonal structure.
Azencot et al. \cite{azencot:2020} present a bidirectional and consistent Koopman autoencoder (CKAE) that includes backward dynamics. The authors argue that a \emph{consistency} loss term moves the eigenvalues of $\mathbf{K}$ close to the unit circle and therefore improves stability, which could not be reproduced in the results of the studies conducted here.

In line with the autoencoder-LSTM models presented in Section \ref{sec:ae_lstm} and similar to the model of Takeishi et al. \cite{takeishi:2017}, we construct a KAE that predicts the next time step $\mathbf{x}_{t+1}$ using the three previous time steps $\mathbf{x}_{t-2}$, $\mathbf{x}_{t-1}$, $\mathbf{x}_{t}$. The decoder uses the same layers as in the CJMDD, cf. Table \ref{tab:KAE}. Note that the three temporally separate outputs of the last convolutional layer in the encoder are first concatenated and afterwards flattened to get a single latent space representation. This procedure could also be applied in the autoencoder-LSTM models. 
Then, information of all time steps is bundled in a single dense layer before entering the LSTM layers. While multiple tests have shown similarly good results of this approach, we keep the time slices separate in the encoder of the 
autoencoder-LSTM models in this work. This is more sensible when multiple latent representations should be the inputs for the LSTM layers.

\begin{table}[h!]
\caption{KAE structure. Conv2D($f$, $k$, $s$) and Conv2DT($f$, $k$, $s$) denote a 2D convolutional layer and 2D transposed convolutional layer, respectively, with $f$ filters of kernel size $k\times k$ and stride $s$ in both dimensions. The shapes do not include the minibatch size. The network is trained on $3$ consecutive grayscale patches of shape $128\times128$, we use channels last, thus the input is of shape $3\times128\times128\times1$. In the convolutional part of the encoder we keep the three time slices separate.}
\centering
\begin{tabular}{cccc}
\hline
\hline
    Layer & Output shape &  Layer & Output shape\\
    \hline
 Input & (3,128,128,1) & Dense(4096) & (4096) \\
     Conv2D(8,3,2) & (3,64,64,8) & Reshape(8,8,64) & (8,8,64) \\
     Conv2D(16,3,2) & (3,32,32,16) & Conv2DT(32,3,2) & (16,16,32) \\
     Conv2D(32,3,2) & (3,16,16,32) & Conv2DT(16,3,2) & (32,32,16) \\
     Conv2D(64,3,2) & (3,8,8,64) & Conv2DT(8,3,2) & (64,64,8) \\
     Flatten() & (12288)  & Conv2DT(1,3,2) & (128,128,1)  \\
     Dense(10) & (10) & Reshape(128,128) & (128,128) \\
     Dense(10) & (10) &  \\
     \hline
     \hline
\end{tabular}
\label{tab:KAE}
\end{table}
The function $\mathbf{K}$ is implemented as a linear layer without bias vector and initialized as a block diagonal matrix with $2\times2$ rotations as diagonal blocks,
\begin{align}
    \mathbf{K}_\text{B} = \begin{pmatrix}
        \phantom{+}\cos (\phi) & \sin (\phi) \\
        -\sin (\phi) & \cos (\phi)
    \end{pmatrix}, \quad \phi\in(-\pi,\pi].
\end{align}
 The angles $\phi$ are chosen such that the eigenvalues are equidistant on the unit circle, similar to the initialization of \textbf{K} by Otto and Rowley \cite{otto:2019}. 

The net is trained on two inputs given by the sequences $\mathbf{x}_{t-2}$, $\mathbf{x}_{t-1}$, $\mathbf{x}_{t}$ and $\mathbf{x}_{t-1}$, $\mathbf{x}_{t}$, $\mathbf{x}_{t+1}$ with loss function
\begin{multline}
            \mathrm{Loss_{KAE}}(\mathbf{x}_{t-2},\mathbf{x}_{t-1},\mathbf{x}_t;\mathbf{x}_{t+1}) = \alpha_{\textsf{reconst}}\cdot \underbrace{\mathrm{MSE}\left(\mathbf{x}_{t}, \mathbf{d}(\mathbf{e}(\mathbf{x}_{t-2},\mathbf{x}_{t-1}, \mathbf{x}_{t}))\right)}_{\text{Reconstruction loss}} \\
            + \alpha_{\textsf{predict}} \cdot \underbrace{\mathrm{MSE}\left(\mathbf{x}_{t+1}, \mathbf{d}(\mathbf{K}(\mathbf{e}(\mathbf{x}_{t-2},\mathbf{x}_{t-1}, \mathbf{x}_{t})))\right)}_{\text{Prediction loss}}
            + \alpha_{\textsf{linear}} \cdot \underbrace{\mathrm{MSE}\left(\mathbf{e}(\mathbf{x}_{t-1},\mathbf{x}_{t}, \mathbf{x}_{t+1}), \mathbf{K}(\mathbf{e}(\mathbf{x}_{t-2},\mathbf{x}_{t-1}, \mathbf{x}_{t}))\right)}_{\text{Linearity loss}}.
\end{multline}
The weights and biases apart from \textbf{K} are initialized using Glorot uniform initialization and zero, respectively. The last layers in the encoder and the decoder are both linear layers, which is standard, as in the work of Otto and Rowley \cite{otto:2019}. The other layers in the encoder and decoder use \textsf{LeakyReLU} with slope parameter $\alpha=0.01$.  
The training procedure is the same as in the autoencoder-LSTM models, only the loss function differs.

Apart from an LSTM and the Koopman-based approach in the latent space, other models like, e.g., a dense neural network should also be able to provide reasonable results. In this work, we focus on the two former approaches and, with respect to the LSTM-based models, on differences in the encoder-decoder structure, i.e., with or without non-local blocks in the encoder, and with or without a deep decoder after the LSTM.

\section{Test Case and Data Processing}
\label{subsec:data}

\subsection{Fuselage Geometry} 
\begin{figure}[h!]
        \centering
       \includegraphics[width=10cm]{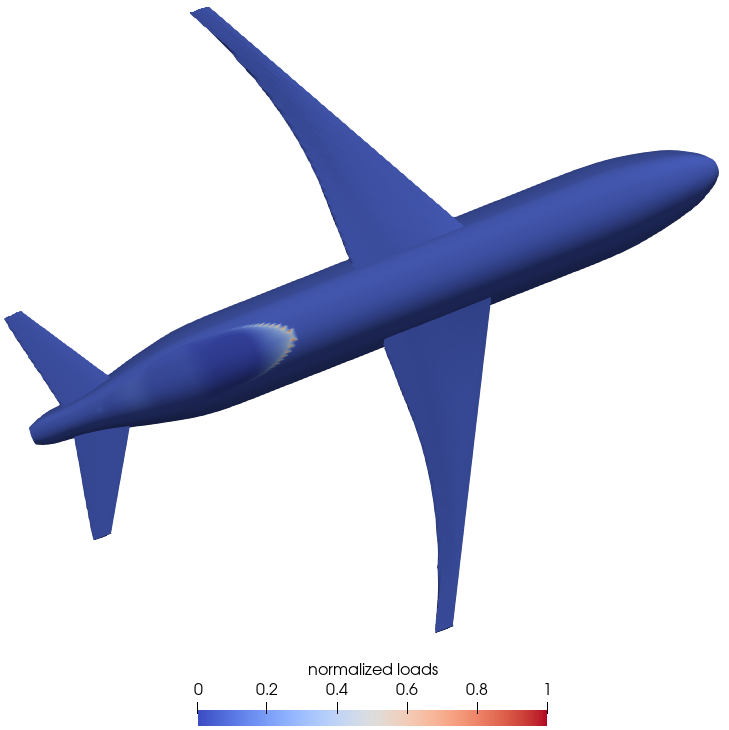}
      \caption{D150 aircraft with exemplary normalized fuselage pressure load contours.}
        \label{fig:aircraft_full}
\end{figure}

The investigated geometry refers to the fuselage of a generic DLR-D150 aircraft, a design that is similar to an Airbus A320 in size \cite{dlr136615}. The length of the aircraft reads $\SI{37.25}{m}$, the span between the wings refers to $\SI{37.07}{m}$. 
Simulations were performed in full scale for the  fuselage-only configuration with a mass of $\SI{68}{t}$ and a center of gravity located at a distance of $\SI{16.43}{m}$ from the nose and vertical height of $\SI{1.72}{m}$ above the bottom of the fuselage.
Though \emph{ditch} is equipped with a simplified aerodynamic and engine/thrust modules, the present study refers to zero thrust and aerodynamics forces, since no wings and engines were considered.    
Fig. \ref{fig:aircraft_full} displays a perspective bottom view of the aircraft including wings and the horizontal tail, which were not considered during the current ditching simulations, and reveals an example of a normalized, instantaneous pressure contour on the fuselage during ditching.

\subsection{Spatial Discretization of the Simulation and Training Data} 
\begin{figure}[h!]
        \centering
        \includegraphics[width=12cm]{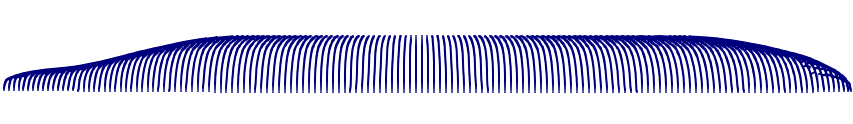}
        \caption{Fuselage frames representing the longitudinal discretization of the ditching simulations.}
        \label{fig:aircraft}
\end{figure}

To prepare the 2D+t simulation, the geometry of each component is discretized in longitudinal ($x$) and cross-sectional ($s$, circumferential) direction. 
Current results were obtained from  $150$ equidistant frames in longitudinal direction and $171$ points along the circumferential direction of each frame
of the investigated fuselage geometry. Mind that \emph{ditch} does not consider roll motion and, therefore, the problem is inherently symmetric. Moreover, the upper part of the fuselage is usually not water wetted, hence the circumferential discretization spans only approximately a bottom quarter of the fuselage at equidistant arc lengths.  
Fig. \ref{fig:aircraft} depicts a bottom view of the discrete frames, where 
the aircraft moves from left to right. 
As parts of the geometry remain unwetted and are not subjected to  hydrodynamic forces during the ditching simulation, segments of $128\times128 = 16 \, 384$ spatial points were selected for the training of the networks, which is only a fraction of the $25 \, 650$ available discrete spatial points.  

\subsection{Temporal Discretization of the Simulation and the Training Data}
The  period of time simulated with the \emph{ditch}-tool covers approximately $T=30$ seconds for the approach, impact and landing phase using 
equidistant time steps of $\Delta t = 5 \times 10^{-3}$ seconds. 
 As the hydrodynamic forces are not acting during the approach phase, the period under consideration for training begins with the time of impact on the free surface --which depends on the approach condition-- and extends over a subsequent, load-intensive phase which lasts a maximum of about 7 seconds for the current investigations. 
  The training is performed with an incremented equidistant training time step 
  of 0.1 second, resulting in at most 70 training time steps. However, only 18-35 time steps were essentially used for training in the majority of cases, since the loads subsequently fell below a lower threshold.
  
Mind that the increment is carried out by using the instantaneous loads in every 20th time step and not through averaging over 20 time steps.

\subsection{Investigated Ditching Scenarios}
The training set consists of data extracted from $323$ different \emph{ditch}-simulations and consists of $8 \, 510$ time-step images in total. Simulations performed for this differed in the vertical and horizontal speed of the aircraft during the approach phase. Considered horizontal velocities range from $\SI{66.88}{m/s}$ to $\SI{87.46}{m/s}$, and vertical velocities in the range from $\SI{0.61}{m/s}$ to $\SI{3.96}{m/s}$ were used. The initial pitch angle was kept fixed at $\SI{6}{\degree}$ for all simulations and only calm-water conditions, 
 assuming zero water current, were investigated.

\begin{figure}[h!]
    \centering
    \includegraphics[width=14cm]{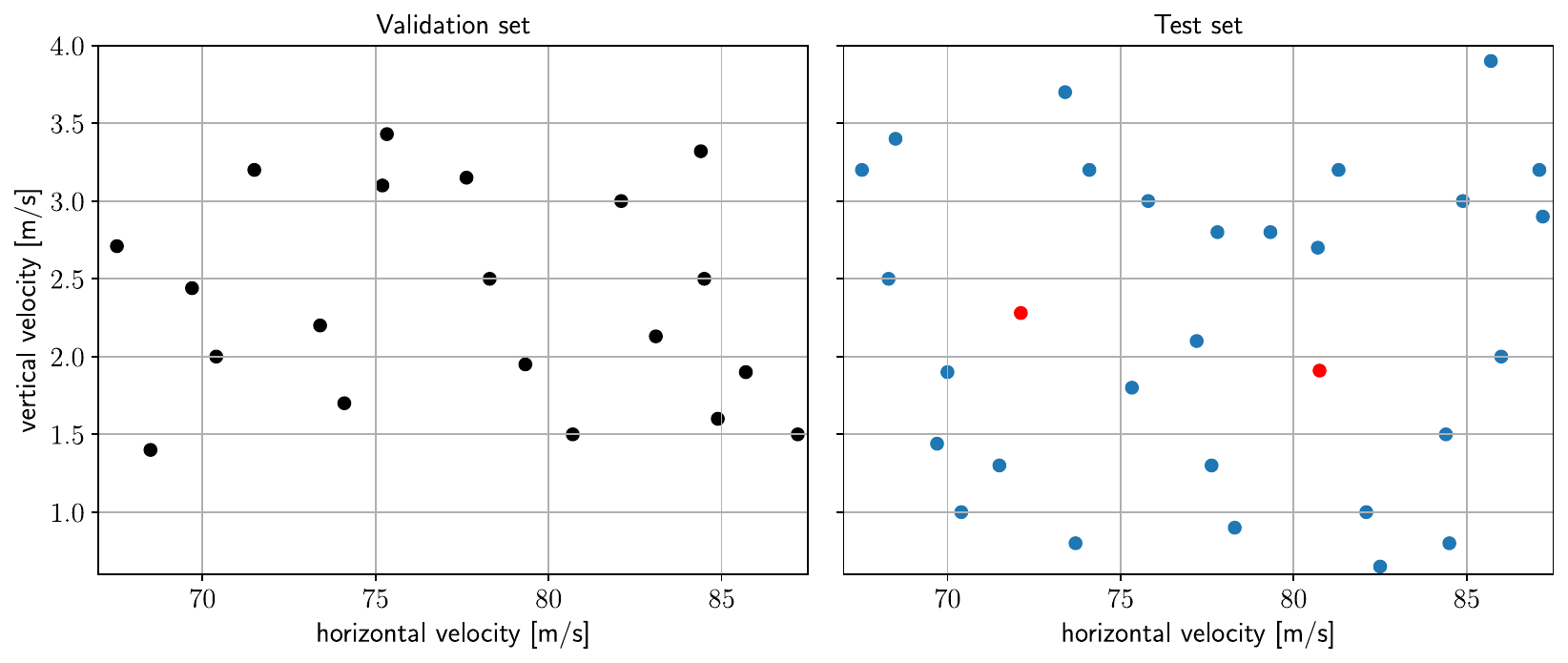}
    \caption{Velocity pairs employed for the validation set (left) and  the test set (right). Red dots refer to the two cases that are analyzed in greater detail.}
    \label{fig:testingset_parameters}
\end{figure}

The validation set involves 20 additional simulations. They consist of 494 time-step images featuring horizontal and vertical velocities in the range of $\SI{67.53}{m/s}$ to $\SI{87.20}{m/s}$ and $\SI{1.40}{m/s}$ to $\SI{3.43}{m/s}$, respectively. Note that these velocities are included in the operating range of the training set. 
The test set comprises data extracted from 30 additional simulations. It consists of 802 time-step images in total and is used to assess the predictive quality of the surrogate models. The related velocity pairs of the validation set and the test set are depicted in Fig. \ref{fig:testingset_parameters}. They are inside the operating range used for training. Two of these cases are analyzed in detail and marked in red in Fig. \ref{fig:testingset_parameters}. The \textbf{first case} refers to a horizontal and vertical velocity of $\SI{72.12}{m/s}$ and $\SI{2.28}{m/s}$. The \textbf{second case} employs a horizontal velocity of 
 $\SI{80.75}{m/s}$ and a vertical velocity of $\SI{1.91}{m/s}$. 

 The two cases differ in the temporal load development, where the maximum pressure loads occur at $t=15$ for the first case and $t=7$ for the second case. Additionally, a larger area of the fuselage is exposed to hydrodynamic pressure forces for the second case, cf. Section \ref{sec:res}.
Note that the first $\ell=3$ time steps of the impact are not taken into account in the above information on maximum values, since they are prescribed to start (initialize) the surrogate model and therefore the predictive quality  is not assessed during the first 3 time steps.

\subsection{Data Processing}
\label{sec:data_processing}
\begin{figure}
    \centering
    \includegraphics[width=15cm]{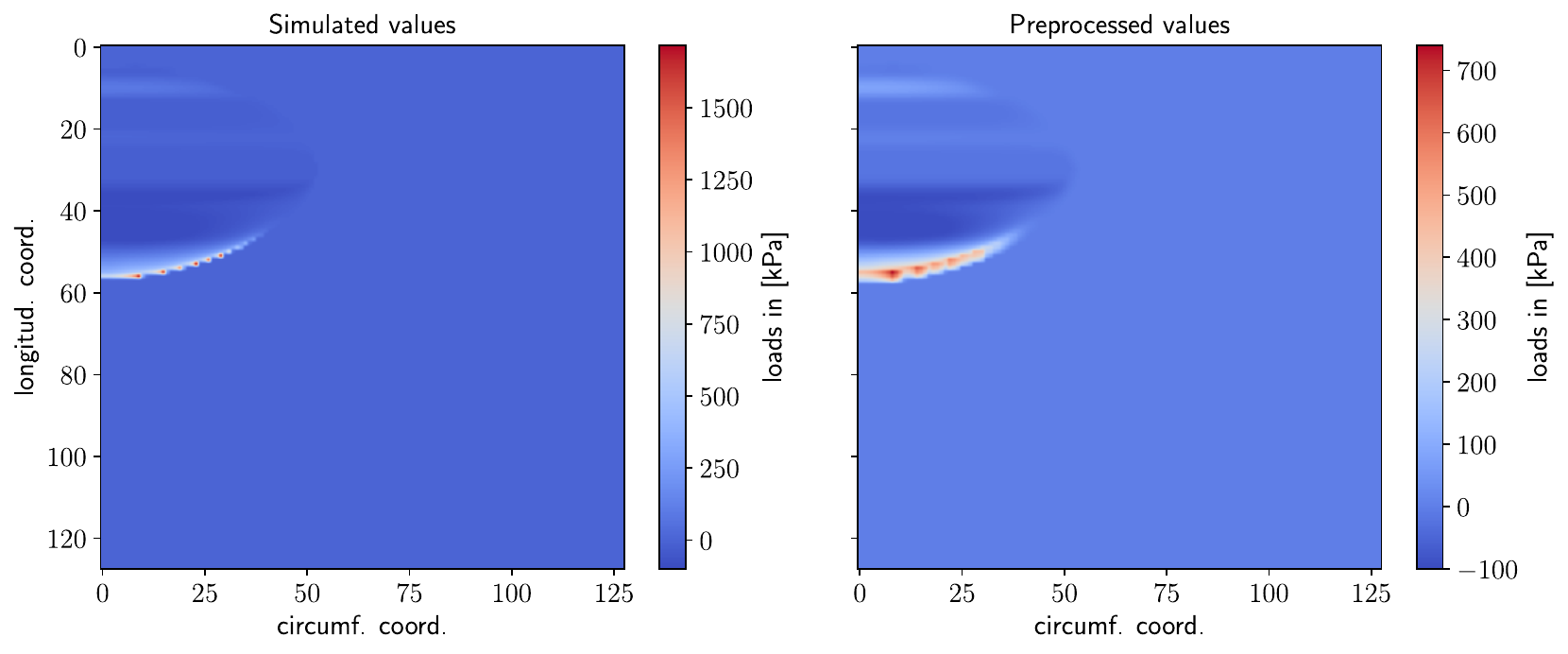}
    \caption{Exemplary pressure force image acting on the D150 fuselage without (left) and with (right) Gaussian blur.}
    \label{fig:blurred_example}
\end{figure}

First tests indicated that the various autoencoder models have severe difficulties to reconstruct load data featuring thin edges that contain isolated peak loads, cf. example on the left side of Fig. \ref{fig:blurred_example}.
The aim of the research is to approximately account for structural deformations under hydrodynamic loads in a two-way coupling using ML. Due to the supporting properties of an intact structure, the load transfer takes place into a spatial area of coarser resolution than the load model when the deformation and not the structural failure is of central interest. In this case, it seems justified to preprocess the data with a Gaussian filter.  
A filter size of $3\times 3$ is the smallest and thus the most reasonable choice. 
The applied filter preserves the total sum of the pressure load, therefore the load signal becomes slightly broadened as indicated by Fig. \ref{fig:blurred_example}. 
After preprocessing, the peak loads of both test cases 
are reduced from $\SI{1720}{kPa}$ to $\SI{740}{kPa}$
for the first case,  
and from $\SI{2370}{kPa}$ to $\SI{810}{kPa}$  
for the second case, which approximately scales with the filter width. Mind that no phase shift is introduced by the spatial filtering.
If it is important to recover the unfiltered loadings, one could use a separate convolutional neural network (CNN) to map the filtered data back to the unfiltered data, see Figs. \ref{fig:unfiltered_allpredictions1} and \ref{fig:unfiltered_allpredictions2} in the appendix.

For the autoencoder-based models, all training, validation and test load data is normalized 
to the unit interval of the training data results  
viz. 
\begin{align}
    \begin{split}
        x_{\max} &= \max(x_{\text{train}}),\qquad 
        x_{\min} = \min(x_{\text{train}}),\\[3mm]
        x_{\text{train}}^{\text{new}} &= \frac{x_{\text{train}} - x_{\min}}{x_{\max}-x_{\min}},\quad 
        x_{\text{val}}^{\text{new}} = \frac{x_{\text{val}} - x_{\min}}{x_{\max}-x_{\min}},\quad 
        x_{\text{test}}^{\text{new}} = \frac{x_{\text{test}} - x_{\min}}{x_{\max}-x_{\min}},
    \end{split}
\end{align}
where the mathematical operations are performed pointwise as $x_{\text{train;val;test}}$ describe the training, validation and test sets respectively. 
The output can be easily transformed back to the original range.

\section{Results}
\label{sec:res}

\subsection{Dynamic Mode Decomposition}
\begin{figure}[h!]
    \centering
    \includegraphics[width=16cm]{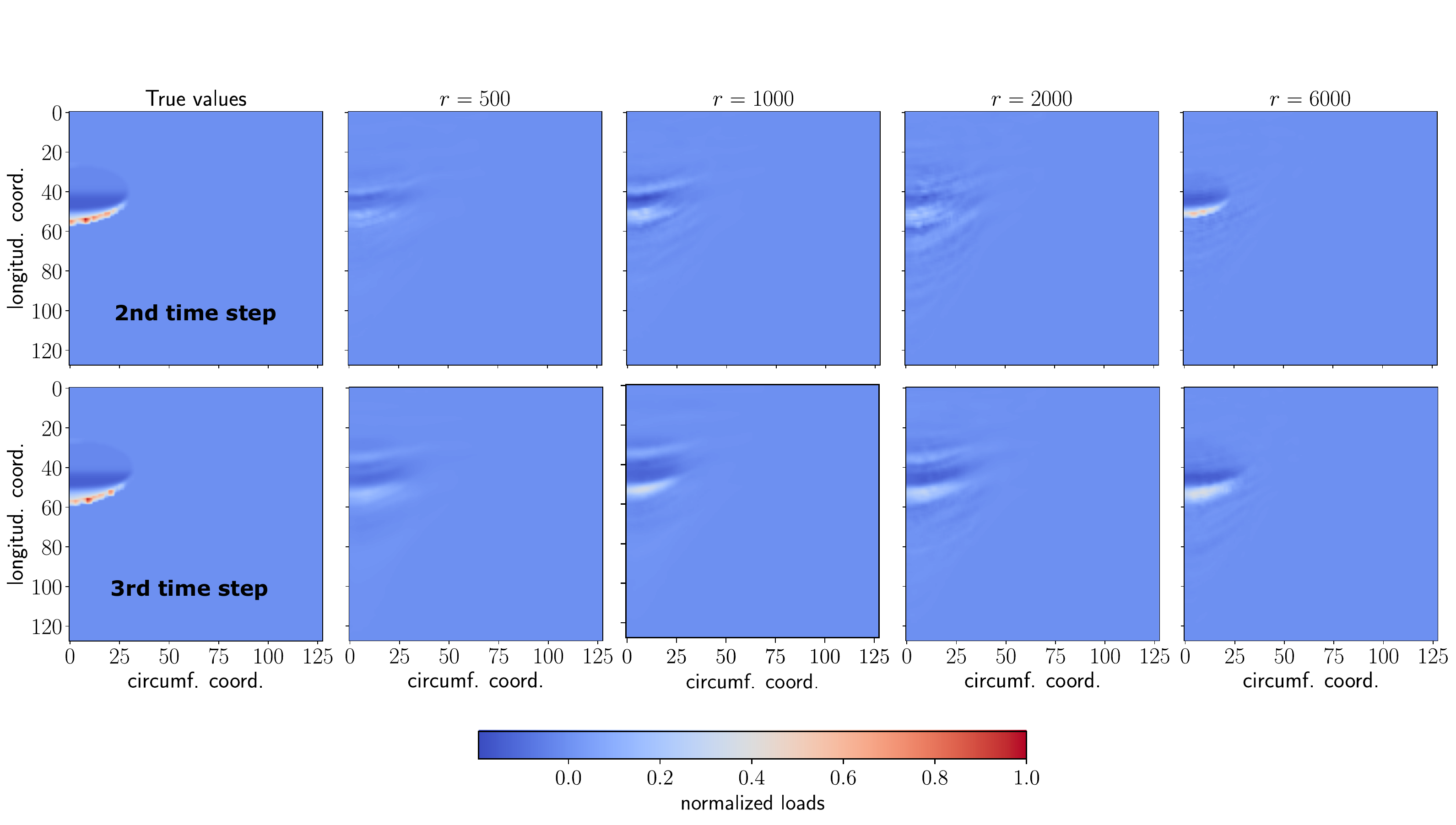}
    \caption{Comparison of simulated  (left) and DMD-based load predictions for the first two time steps of an exemplary test case. Graphs display results obtained with different ranks $r$ of the SVD. Loads are normalized by the maximum load of $\SI{734}{kPa}$. 
    }
    \label{fig:dmd}
\end{figure}

To demonstrate that DMD is sub-optimal for the accurate and fast prediction of ditching loads, we use the Python library PyDMD \cite{pydmd2018, pydmd2024} and perform exact DMD \cite{tu:2014} on the full dimension with four different ranks $r$ of the singular value decomposition (SVD), i.e., $r=500, 1000, 2000, 6000$.
Fig. \ref{fig:dmd} displays the first two predicted time steps and the corresponding true values for an exemplary test case. The initial time step of the impact was used to start the DMD. Using the SVD ranks up to $r=2000$ yields poor predictions that underestimate the peak loads already in the first predicted time step. For $r=6000$ a more reasonable pressure distribution is predicted for the first time step. However, the second prediction again reveals clear deficits. In addition, such high SVD ranks are afflicted by a large  computational effort: For $r=6000$, predicting only one time step requires about 85-100 seconds on an Intel Xeon Silver 4314 Processor, which renders the method unfavourable in this work.

\subsection{Model Comparison}
Table \ref{tab:model_comparison} summarizes the number of trainable parameters, the approximate training time and the average inference time experienced on a workstation with a NVIDIA GeForce RTX 3090 GPU. All autoencoder-LSTM models are trained for 500 epochs. The KAE is trained for 150 epochs to reach the same training quality.
Most of the parameters of the CJM and the CJMNLB are used for the LSTM network or, more precisely, the dense layer that follows the LSTM layers. The CJMDD has less parameters, which is due to the fact that it uses a lot less dense connections after the LSTM layers. The KAE holds the least parameters as there is only one linear layer inside the latent space compared to two LSTM layers.
With around 0.04-0.07s per prediction, the inference time is low for all investigated models.

\begin{table}[h!]
\caption{Number of trainable parameters, training time and inference time for the tested models.}
\centering
\begin{tabular}{cccc}
    \hline
    \hline
    Model & Trainable parameters& Training time [s] & Time per prediction [s] \\
    \hline
    CJM & 1,844,938 & 430 & 0.041\\
    CJMDD & 260,541 & 490 & 0.045\\
    CJMNLB & 1,855,178 & 600 & 0.049\\
    KAE & 216,751 & 475 & 0.068 \\
    \hline
    \hline
\end{tabular}
\label{tab:model_comparison}
\end{table}

 Suitable values for the loss function coefficients of the KAE, cf. Eq. (\ref{eq:kaeloss}),  were identified by  $\alpha_{\textsf{reconst}}=\alpha_{\textsf{predict}}=1$ and $\alpha_{\textsf{linear}}=0.01$. For larger values of $\alpha_{\textsf{linear}}$, the linearity loss term is too restrictive and causes the training to saturate at
  poor minima or saddle points. In such cases, an undesirable  trivial solution is found, in which the encoder maps every input to nearly zero. This issue may  sometimes even occur  
  in conjunction with $\alpha_{\textsf{linear}}=0.01$. 
  Our studies indicate a way out by ignoring the linearity loss term at the beginning of the training. This results in a more favorable training process and is suggested similarly by Lusch et al. \cite{lusch:2018}. 
  To this end, 
 the linearity loss term is not considered 
 during the initial training phase with $\gamma=25$ epochs. The model thereby first identifies suitable parameters for the reconstruction and prediction losses during the initial training phase, and the training is no longer attracted towards the trivial solution.

\subsection{Load Prediction}
\label{subsec:load_prediction}
First, the models are tested on the two test cases described in Section \ref{subsec:data} and analyzed in detail. The cases consist of 27 and 24 time steps, respectively, leading to 24 and 21 time steps that are assessed  following the first 3 specified time steps. As error measure, the \emph{root mean squared error} (RMSE), which is defined as the square root of the MSE, is employed.
To account for the randomness in the training process, each model is trained five times, and 
predictions for both cases are obtained five times for each model. Afterwards, the averages, best and worst of the five obtained RMSE vectors for each model are computed for both cases.
We show the KAE predictions obtained by the full model, i.e., the input sequence $\mathbf{x}_{t-2},\mathbf{x}_{t-1}, \mathbf{x}_t$ is encoded, then one prediction is performed inside the latent space and subsequently decoded to get $\mathbf{x}_{t+1}$. The sequence is updated to $\mathbf{x}_{t-1},\mathbf{x}_{t}, \mathbf{x}_{t+1}$ and the procedure is repeated for the new sequence.
In our experiments, this approach lead to consistently better performance compared to solely predicting in the latent space without mapping to the full dimension in each time step. Figs. \ref{fig:kae_latent_predictions1} and \ref{fig:kae_latent_predictions2} of the appendix outline the superior results of the current KAE approach over the latent space predictions, which also meets with the findings reported for EDMD dictionary learning \cite{Li:2017} approaches in \cite{Constante-Amores:2024}.  However, we note that the dynamics are then no longer modelled linearly due to the employment of the nonlinear decoder and encoder between the predictions, as already stated in \cite{Constante-Amores:2024} for the EDMD dictionary learning approach.
The approach is consistent with the autoencoder-LSTM models. It is the only possible procedure for those autoencoder-LSTM methods that reconstruct the data directly in the full dimension after the LSTM (CJM, CJMNLB). Using the CJMDD architecture, however, it would also be possible to
perform the time evolution only in the latent space. To this end, the loss function
would have to include loss terms for the reconstruction and the temporal evolution
in the latent space. We believe this makes training the model more difficult, since
the two loss terms have to be balanced analogously to the KAE approach.

Secondly, to obtain more general information about the errors resulting from the different models, all 30 test cases described in Section \ref{subsec:data} are considered.

\subsubsection{Integral Error Measures}
\begin{figure}[h!]
    \centering
    \includegraphics[width=11cm]{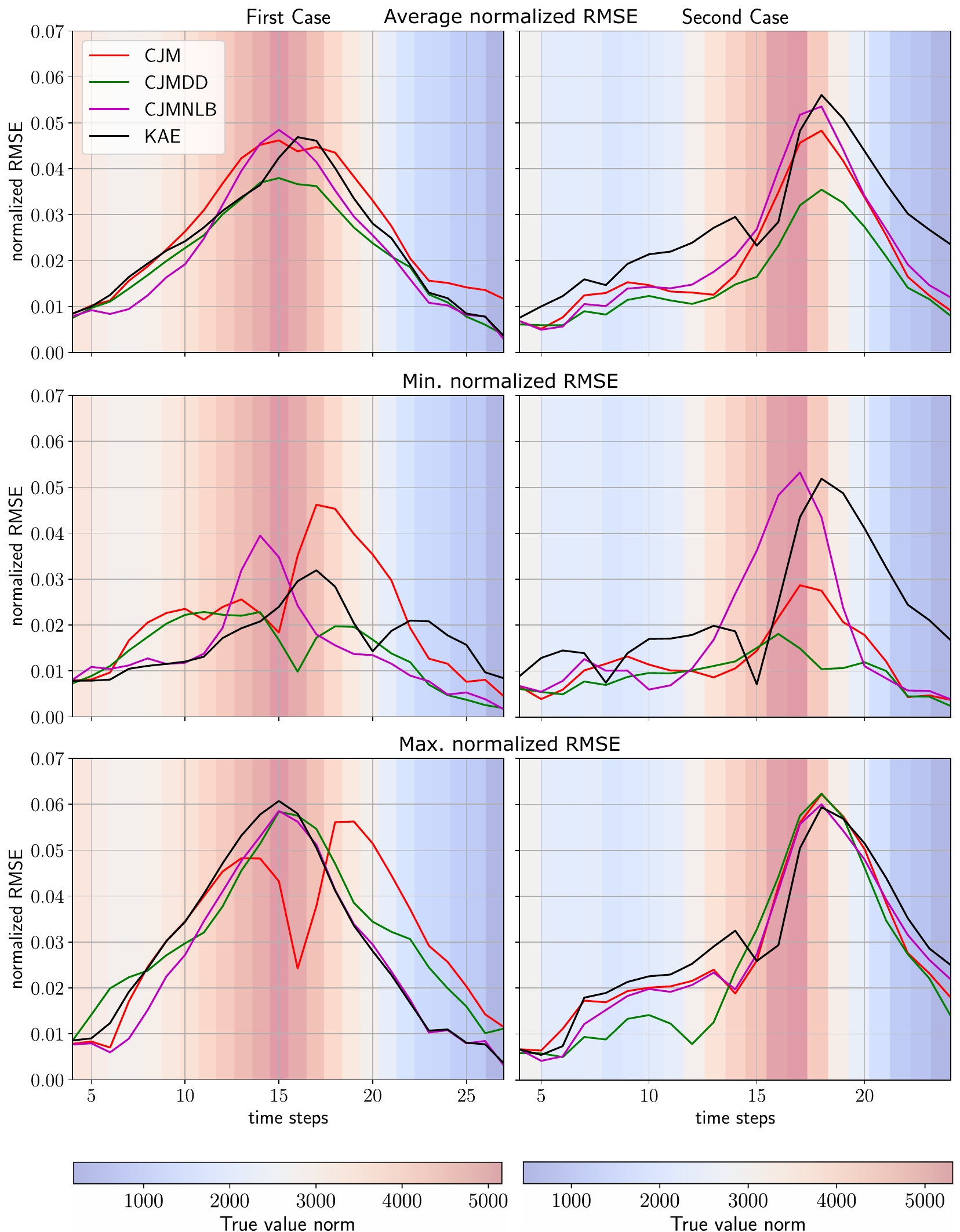}
    \caption{Temporal evolution of the average (top), minimum (center) and maximum (bottom) observed normalized RMSE of the four investigated surrogate models 
    for both test cases. Respective averages were extracted for five different trainings.
    Background color refers to the norm of the true value to indicate the relative error.} 
    \label{fig:rmse}
\end{figure}

Fig. \ref{fig:rmse} depicts the average RMSEs (top), as well as the best (center) and worst (bottom) RMSEs, each normalized by the maximum load of the respective case. The figure distinguishes between case 1 (left) and case 2 (right).   
In addition, the Frobenius norm of the true simulated loads is illustrated in the background for each time step to indicate the qualitative error behaviour. Mind that the first $\ell=3$ time steps after the impact were used to start the surrogate models. 

Considering the first case, represented by the left graphs of Fig. \ref{fig:rmse}, the average errors of the investigated models are generally close to each other. The average lowest RMSE value is returned by the CJMDD approach. The average RMSE performance of the  other three surrogate models lags behind the CJMDD results, with the KAE performing slightly better than the CJM and CJMNLB.
This finding also applies for the minimum RMSE results.   
For all LSTM-based models, 
the largest average errors occur at the 15th time step of case 1, where the norm of the true simulated loads has its maximum over all time steps. The average RMSE of the KAE however peaks slightly delayed. 
Both, the true values as well as the average errors decrease for $t>15$, hence the average predictions 
are virtually in phase with the true values. However,  
minor phase issues occur for the CJM approach in conjunction with the minimum and maximum RMSEs.
Looking at the minimum RMSE evolution, the best performance for the first case is achieved by the CJMDD and the KAE approaches. At any time, the errors of both models remain smaller than 
about 2.3\% and 3.2\% of the peak load, respectively. 
On the contrary, the CJM and the CJMNLB yield significantly larger  minimal RMSE values, which increase by approximately a factor of 2 as compared to the CJMDD. 

With attention directed to the second case, displayed in the right column of Fig. \ref{fig:rmse}, 
the CJMDD again performs superior and reveals significant benefits over the other three surrogate models, and average RMSE values of the CJMDD predictions are approximately 35\% smaller.
The predictions of the KAE depict the largest average errors during both, the initial 11 time steps and the final 5 time steps. 
In contrast to the first case, the predicted average RMSE peaks of all surrogate models are slightly retarded from the peak of the true value.
As regards the minimum RMSE evolution for the second case,
the errors displayed by the CJMDD and the CJM are smaller than 2\% and 3\% of the peak loads, respectively, for all time steps. During the first 12 time steps, the errors of the KAE remain smaller than 2\% of the peak loads, but increase up to about 5\% 
before decreasing again. The CJMNLB shows a similar error development.

\smallskip
Focusing upon the average errors returned by the KAE, the model performs similarly well as the CJMNLB and the CJM in the first case, and even better looking at the minimum RMSEs. For the second case, on the other hand, the KAE falls short of these two poorer performing LSTM models.

\subsubsection{Exemplary Spatio-Temporal Load Distributions}
Figs. \ref{fig:allpredictions1} and  \ref{fig:allpredictions2} display the pressure load predictions for 5 selected time steps of the first and the second case, respectively. Results are obtained from best predictions for all models and correspond to the lowest RMSEs shown in Fig. \ref{fig:rmse}. The selected time steps include the peak loads 
at time step 15 and 17 for the first and second case, respectively. Again, all pressure loads are normalized by the corresponding maximum load. 

We can observe that the predictions are generally in satisfactory agreement with the true values displayed in the respective top row. 
Focusing upon the first case indicated by Fig. \ref{fig:allpredictions1}, the true load pattern can easily be recognized in most predictions. This finding even holds for the last depicted time step (right column), when the maximum of the instantaneous pressure forces are already nearly an order of magnitude lower than before. At this time step, only the KAE prediction significantly exceeds the true peak loads.
All models yield satisfactory predictions of the peak loads at the 15th time step. However, we observe small shifts of the peak pressure loads predicted by the CJMNLB (forward) and the KAE (rearward), as marked by the dotted lines. 
At the 18th time step, the  peak pressure load predicted by the CJM is smaller compared to the true peak load and the prediction of the other models. Furthermore, the predicted peak of the CJM at this time step is also exposed to a small rearward shift. Although the predicted load pattern of the CJM seems  acceptable, the occurrence of small shifts explains the augmented error levels observed in Fig. \ref{fig:rmse}. In contrast to this, the location and the level of the CJMDD peak predictions are quite accurate at every time step, which can be expected from the small error levels seen in Fig. \ref{fig:rmse}. 

Fig. \ref{fig:allpredictions2} outlines the analogue results for the second case. We observe high quality predictions until the 13th time step, which is also apparent in Fig. \ref{fig:rmse}. 
At the 17th time step, the CJMNLB clearly performs worse than the other models and displays a rearward shift of its slightly small peak. The CJM and the CJMDD show satisfactory predictions of the peak load levels at this time step. Moreover, we observe a small forward shift of the predicted peak loads by the KAE, as well as an even smaller rearward shift of the CJM predicted peak, in line with the increased errors documented in Fig. \ref{fig:rmse}.
The subsequent pressure load drop is correctly predicted by all models. However, there are locations, in which the KAE predictions exceed the true loads, as suspected by the delayed error decrease in Fig. \ref{fig:rmse}.

We also note small but visible oscillations in the CJM and CJMNLB predictions for both cases, which are more clearly observed in Fig. \ref{fig:8thTimestep}, which refers to the 8th time step of case 1. Different from the other two models, both of these models do not use a deep convolutional decoder, which indicates an improved performance of the second LSTM variant (II). 
Due to the significantly higher number of parameters in the CJM and CJMNLB, this could seem to be related to overfitting. But this is not the case here as the oscillations are also visible when the models are tested on training data.
In fact, we report that the oscillations also occur when reconstructing the loads using a proper orthogonal decomposition (POD) \cite{lumley1967structure} with 100 modes, cf. Fig. \ref{fig:pod_reconstruction} in the appendix. Due to the linearity of the last layer of the CJM and the CJMNLB, the lowest possible reconstruction error in steady state, i.e., without time evolution,  that could be achieved with this architectures would be the reconstruction error obtained from a POD with 100 modes. Though the reconstruction errors of the POD, which feature an average normalized RMSE of 0.006, are lower than the prediction errors of the CJM and CJMNLB and the linearity of the last layer therefore does not lead to higher errors by construction,  100 modes seem to be too few to capture the high frequency content (cf. also comment in Subsection \ref{sec:ae_lstm}).
Furthermore, including non-local blocks into the CJM does not yield significantly better results for the two considered cases. Both models reach very similar average errors on both test cases.
The CJMNLB achieves more accurate predictions for the first case, and for the second case, the CJM performs better. 

\subsubsection{Transient Peak Loads}
The accurate prediction of the peak load instant is of significance for the correlated load prediction \cite{lazzara:2022}. 
Fig. \ref{fig:maxima_times} outlines an assessment of the peak load time instant for the best performing models, corresponding to the data already displayed in Figs. \ref{fig:allpredictions1} and \ref{fig:allpredictions2}.
The figure reveals deficits of using the 
CJM and CJMNLB models 
in regimes of vanishing loads. On the contrary, the CJMDD and the KAE display an improved prediction of the load history for the first case. 
For the second case, the CJMDD clearly reveals the most accurate prediction of the load history, while the KAE fails to cover the occurrence of pressure forces  towards the front of the aircraft. 

Apart from the better performance of the CJMDD, for which the errors remain small  for all time steps, we observe a clear pattern in the error development: For both investigated cases, the errors of all models increase when the true values increase. 
Even if the true values are not predicted accurately in some time steps, this implies that the models learn the underlying load development 
 which is also apparent in Fig. \ref{fig:maxima_times}. Moreover, a higher error can also be the result of a small spatio/temporal shift of the peak loads, as seen in Figs. \ref{fig:allpredictions1} and \ref{fig:allpredictions2}.

 \subsubsection{Global Comparison}
For the global error comparison, predictions are again performed with all five trained versions of each model and the normalized RMSEs are computed for each test case. We then average these normalized RMSEs along the time separately for the 30 test cases, compare the performance of all 5 versions and determine the related best practice. Subsequently, we compare the four best practices obtained for the four models and denote the overall best practice for each test case, cf. Table \ref{tab:model_error_comparison}.
Moreover, we calculate the \emph{total average error} for each model. To this end, we compute the RMSEs of one model and average these for its five versions on a particular test case. We subsequently average this quantity in time for each test case, and finally average the resulting 30 numbers over all test cases, cf. Table \ref{tab:model_error_comparison}.

\begin{table}[h!]
\caption{Number of times, the lowest temporally averaged RMSE is reached by a model on a test case and total average errors obtained on all test cases.}
\centering
\begin{tabular}{cccc}
    \hline
    \hline
    Model & Total number & Percentage [\%] & Total average error\\
    \hline
    CJM & 4 &13.33 & 0.024 \\
    CJMDD & 18 & 60 & 0.020 \\
    CJMNLB & 7 & 23.33 & 0.022 \\
    KAE & 1 & 3.33 &  0.025\\
    \hline
    \hline
\end{tabular}
\label{tab:model_error_comparison}
\end{table}
Results in Table \ref{tab:model_error_comparison} confirm the CJMDD to be the best performing model. It yields the lowest total error averaged over the considered 30 test cases and the lowest error in 18 of the 30 test cases, corresponding to 60\%. While the CJM and the CJMNLB show a quite similar average performance on the two test cases discussed in greater detail above, we observe that the CJMNLB achieves the lowest error on 7 of the 30 test cases compared to 4 test cases for the CJM. Additionally, a reduction of the total average error of nearly 10\% compared to the CJM is returned by the CJMNLB, indicating an improved predictive performance through the inclusion of non-local blocks.
The KAE falls short of the LSTM-based models and displays the highest total average error, which is just above the CJM. However, the KAE still manages to return the best performance in one test case.

Fig. \ref{fig:test_set_cjmdd_coloring} depicts the velocity pairs of the test set, colored by the average normalized RMSEs obtained from the five trained versions of the CJMDD. From this figure, a trend can be observed: Ditching events with lower vertical velocities seem to be harder to predict for a surrogate model. We believe that this might be attributed to the more time steps involved in these cases. For the horizontal velocity,  no obvious trend is noticeable.

When the errors are computed after applying a CNN to map the filtered loads back to the unfiltered loads, as mentioned in Subsection \ref{sec:data_processing} and the appendix, the error-based ranking of the models in Table \ref{tab:model_error_comparison} remains the same.

\subsubsection{Increasing the filter size}
The above discussed results employ load data that was blurred with a filter of size $3\times 3$. 
By increasing the filter size to $7\times 7$, the regime containing the peak loads gets wider, while the values of the peak loads decrease. 
Without going into the details, we report that the prediction errors generally behave proportional to the peak loads, i.e., they are roughly half as high as in the $3\times 3$ filter case. Interestingly, the relative performance differences of the tested variants in Table \ref{tab:model_error_comparison}  remain unchanged.

\begin{figure}
    \centering
\includegraphics[width=16cm]{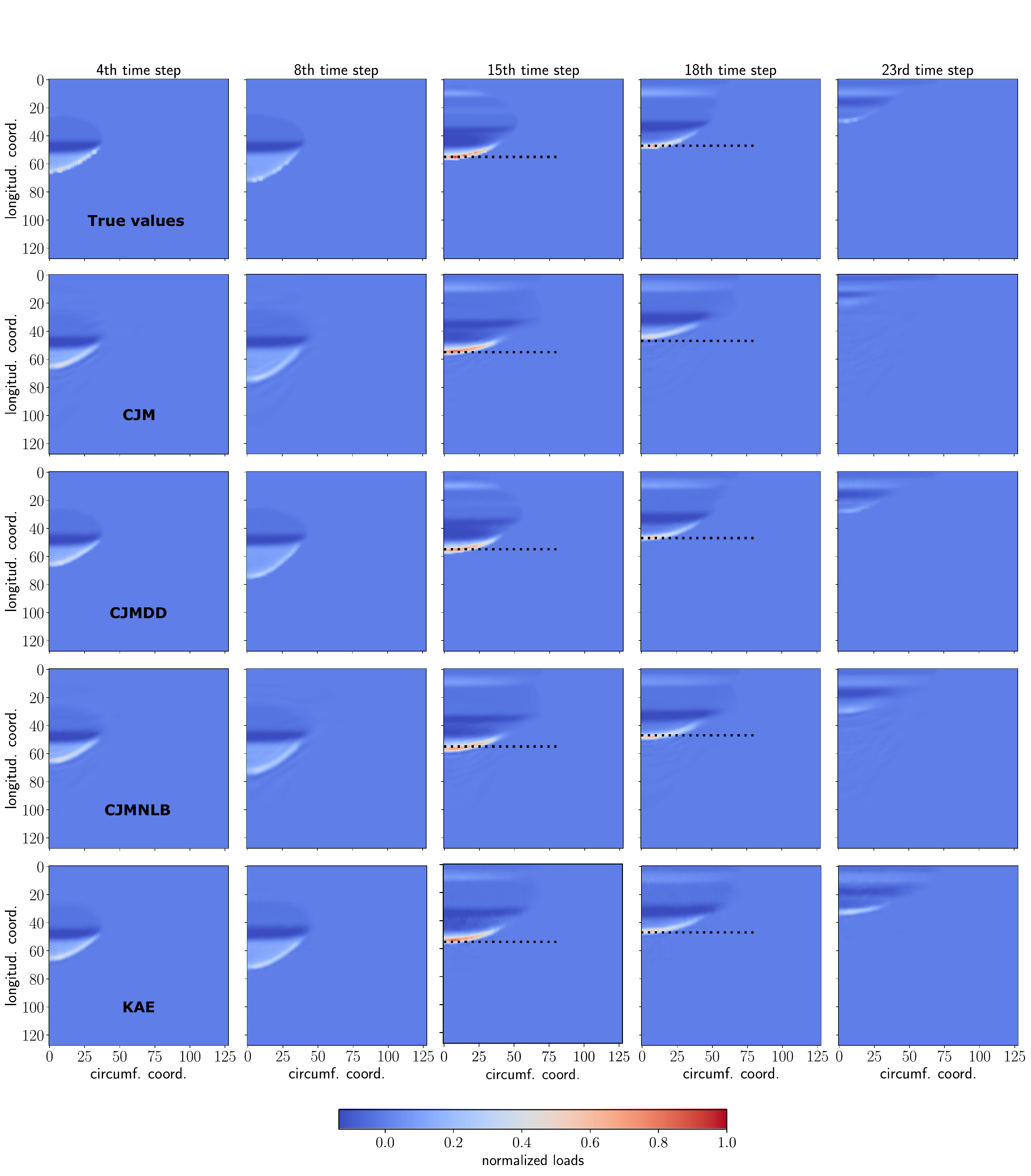}
    \caption{Comparison of numerical  and surrogate model load predictions for selected time steps of the first test case.  Loads are normalized by the maximum load of $\SI{740}{kPa}$. The initial $\ell=3$ time steps were used to start the models. Dotted lines indicate  longitudinal coordinate of true peak load predictions of the current time step.
    }
    \label{fig:allpredictions1}
\end{figure}

\begin{figure}
    \centering
\includegraphics[width=16cm]{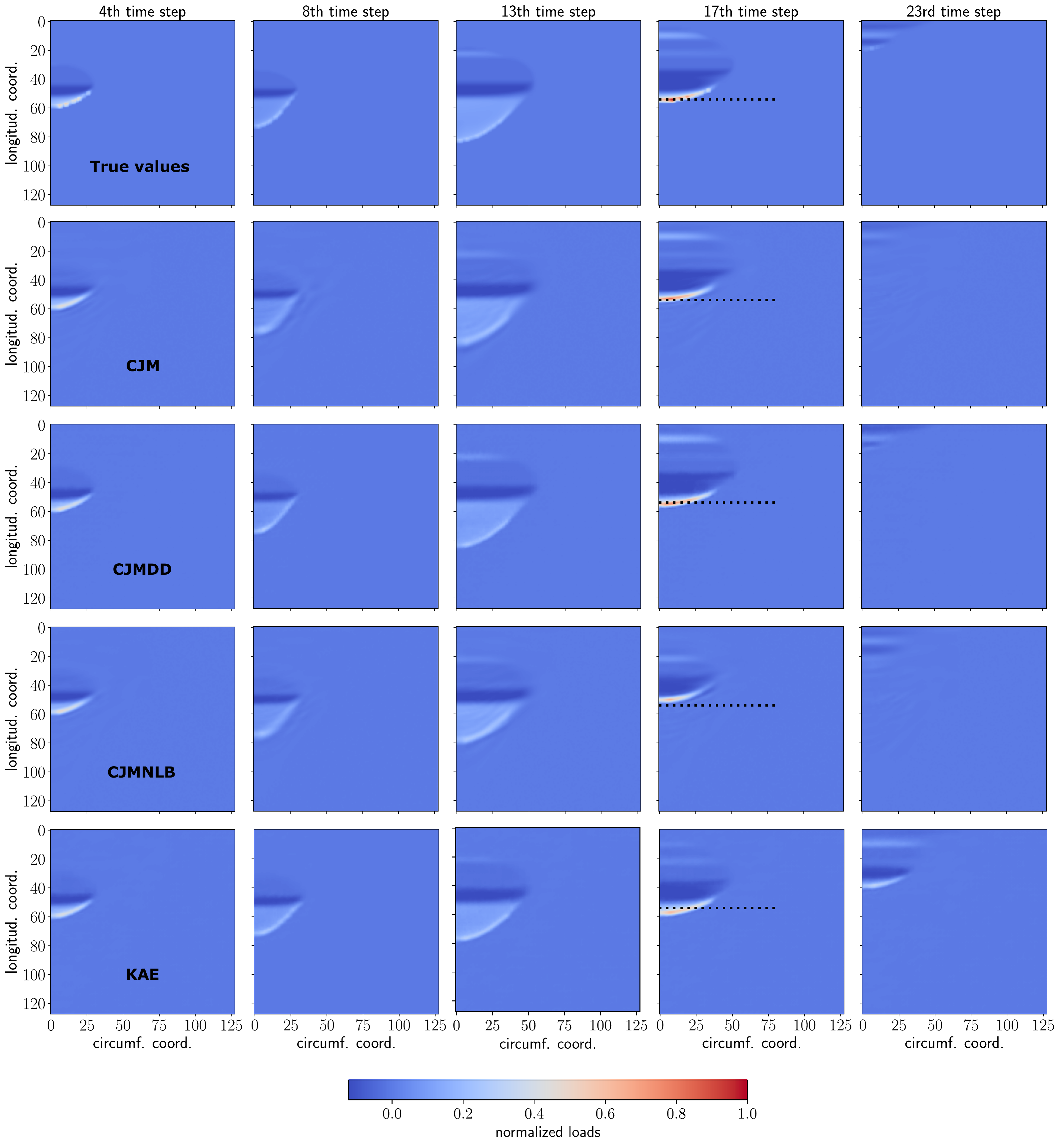}
    \caption{Comparison of numerical and surrogate model  load predictions for selected time steps of the second test case. Loads are normalized by the maximum load of $\SI{810}{kPa}$. The initial $\ell=3$ time steps were used to start the models. Dotted lines cf. Fig \ref{fig:allpredictions1}.
    }
    \label{fig:allpredictions2}
\end{figure}

\begin{figure}
    \centering
    \includegraphics[width=16cm]{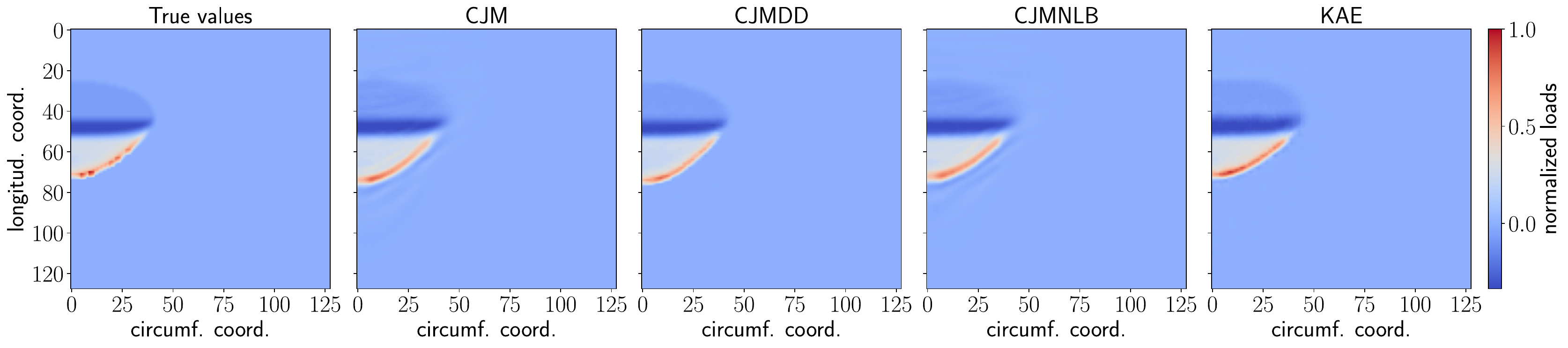}
    \caption{Comparison of numerical
    and surrogate model 
    load predictions for the 8th time step of the first test case.  Loads are normalized by the maximum load of $\SI{298}{kPa}$.}
    \label{fig:8thTimestep}
\end{figure}

\begin{figure}
    \centering
\includegraphics[width=16cm]{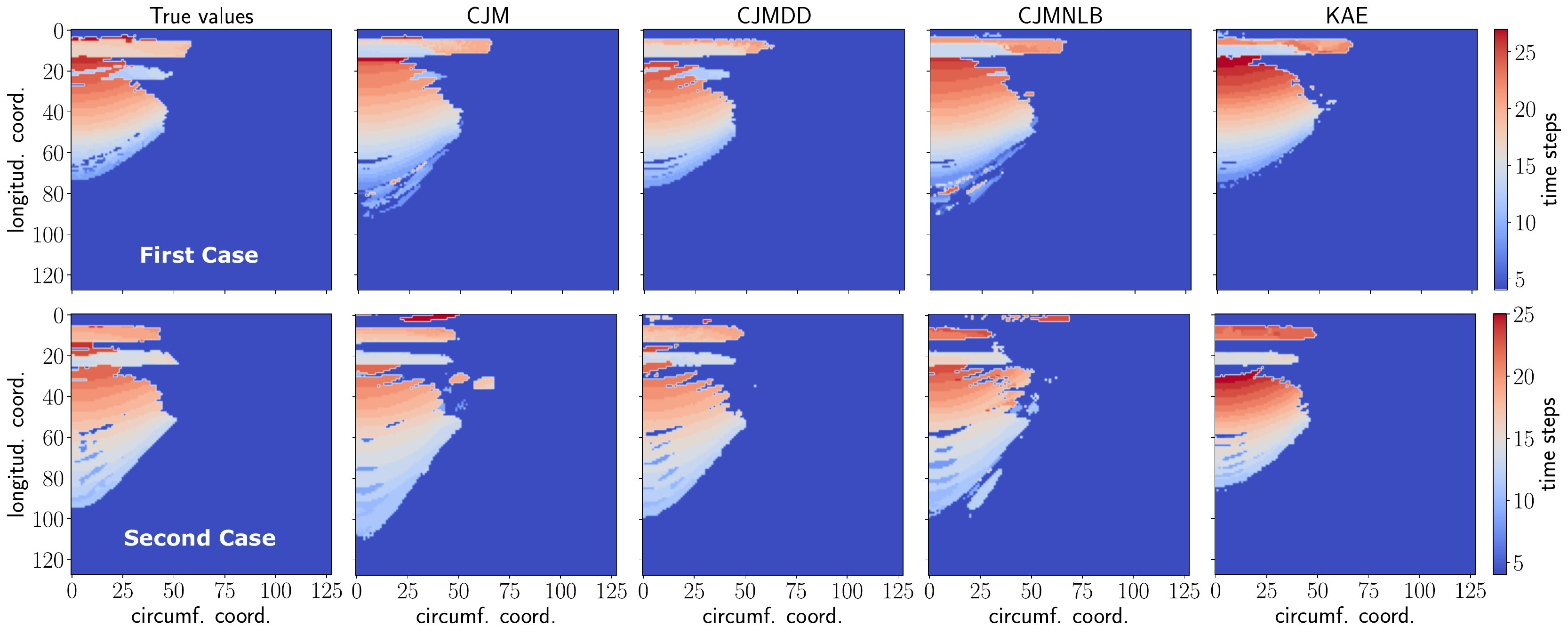}
    \caption{Time instances for the occurrence of maximum values for each point in space. Load values smaller than $\SI{5}{kPa}$ are omitted in the predictions. 
 }
    \label{fig:maxima_times}
\end{figure}

\begin{figure}
    \centering
    \includegraphics[width=0.45\linewidth]{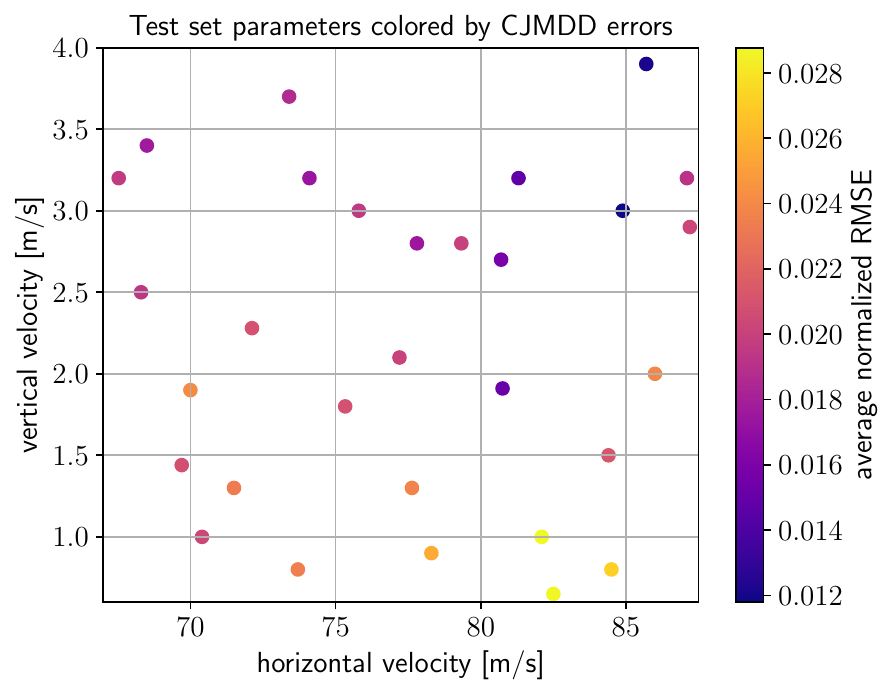}
    \caption{Velocity pairs employed for the test set colored by the average normalized RMSE on each test case obtained from the five trained versions of the CJMDD.}
    \label{fig:test_set_cjmdd_coloring}
\end{figure}

\section{Conclusion}
\label{sec:concl}
The paper reports the performance of different convolutional autoencoder based models for  predicting  ditching load dynamics on an aircraft fuselage using a time-marching ML procedure. 
 Next to spatial load distributions, the temporal accuracy of peak loads was  assessed. 
Results of the present study reveal that the spatial dependencies of the load data are satisfactorily captured by convolutional layers inherent to all investigated models.  
 Moreover, all approaches are associated with negligible costs in the order of 10 milliseconds per time step. Based on full order model runtimes in the tens of seconds or minutes range on an Intel Xeon Silver 4314 CPU, the ML based simulation is approximately two orders of magnitude faster on an NVDIA GeForce RTX 3090 GPU. 
Predictive differences occur when looking at the different options to 
capture the temporal evolution of loads. In this regard, four options were considered, 
using either long short-term memory (LSTM) layers or a Koopman operator approach in the latent space. 
Attention was restricted to joint models, which simultaneously train the global network composed from the dimensionality reduction and the transient part.  
Amongst the three investigated LSTM-based models, the best performance was obtained when using a deep decoder (CJMDD) which --- different from the two investigated LSTM-based models that directly reconstruct the full order data (CJM, CJMNLB) --- revealed no spatial fluctuations 
of predicted loads. 
Moreover, the average errors obtained by the CJMDD were lower than for 
the other models, and a 
satisfactory performance was also achieved by the CJMDD in the prediction of the peak-load time instant. 
The second best performance was obtained by the CJMNLB that includes non-local blocks, showing an improvement to the same model without the non-local blocks (CJM).
In general, predictions using the KAE are also of good quality, especially during the first portion of time steps.

An important aspect refers to the models' trustworthiness and interpretability, which 
be a challenge in future work. Embedding classic (deterministic) elements within ML-based models, such as Koopman operator methods, might provide an added value here.

\section*{Appendix}
\begin{figure}
     \centering
     \includegraphics[width=16cm]{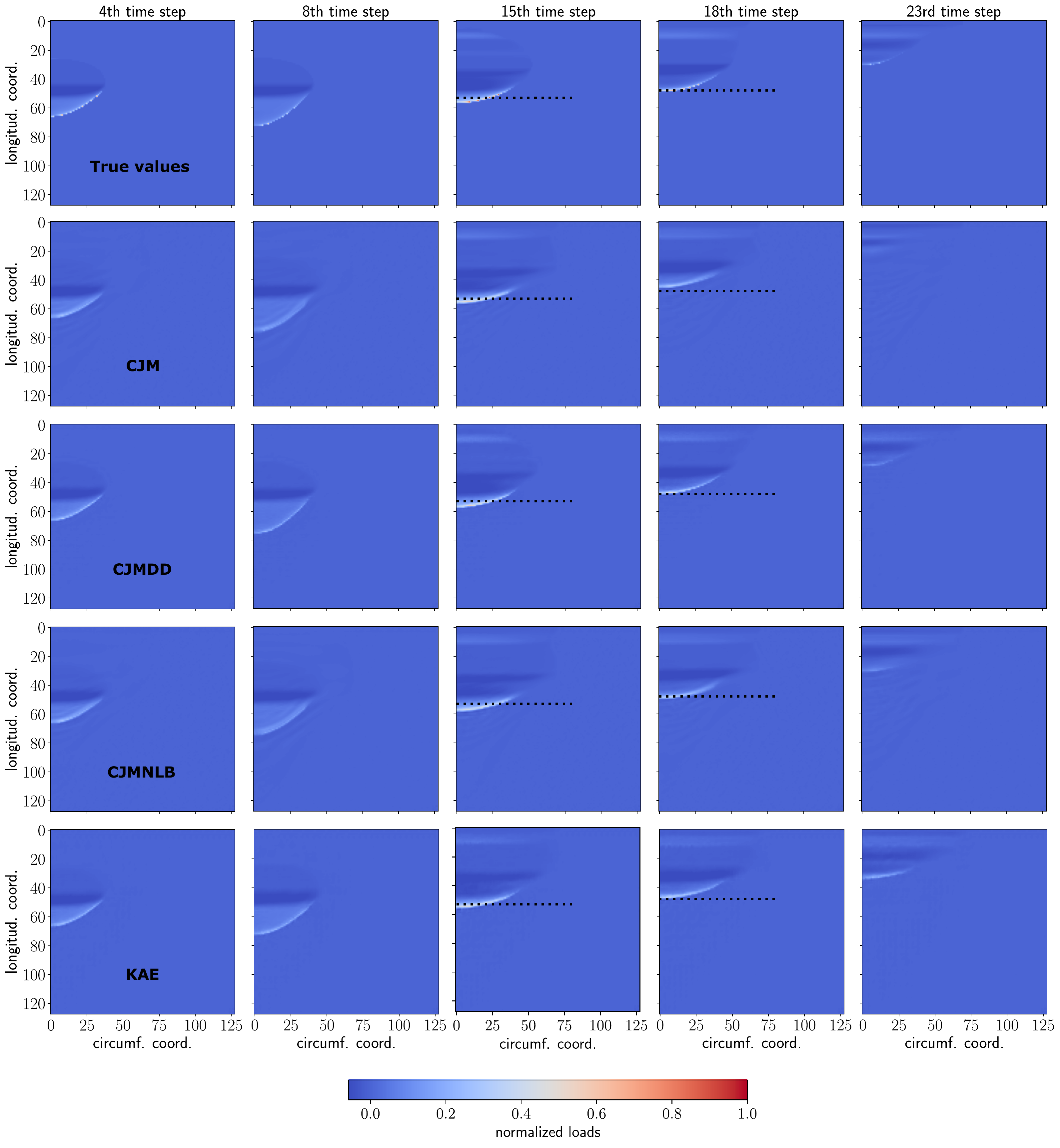}
     \caption{Comparison of unfiltered numerical and surrogate model load predictions for selected time steps (first case). Loads are normalized by the maximum load of $\SI{1720}{kPa}$. The initial $\ell=3$ time steps were used to start the models. Dotted lines cf. Fig \ref{fig:allpredictions1}.}
     \label{fig:unfiltered_allpredictions1}
 \end{figure}

\begin{figure}
     \centering
     \includegraphics[width=16cm]{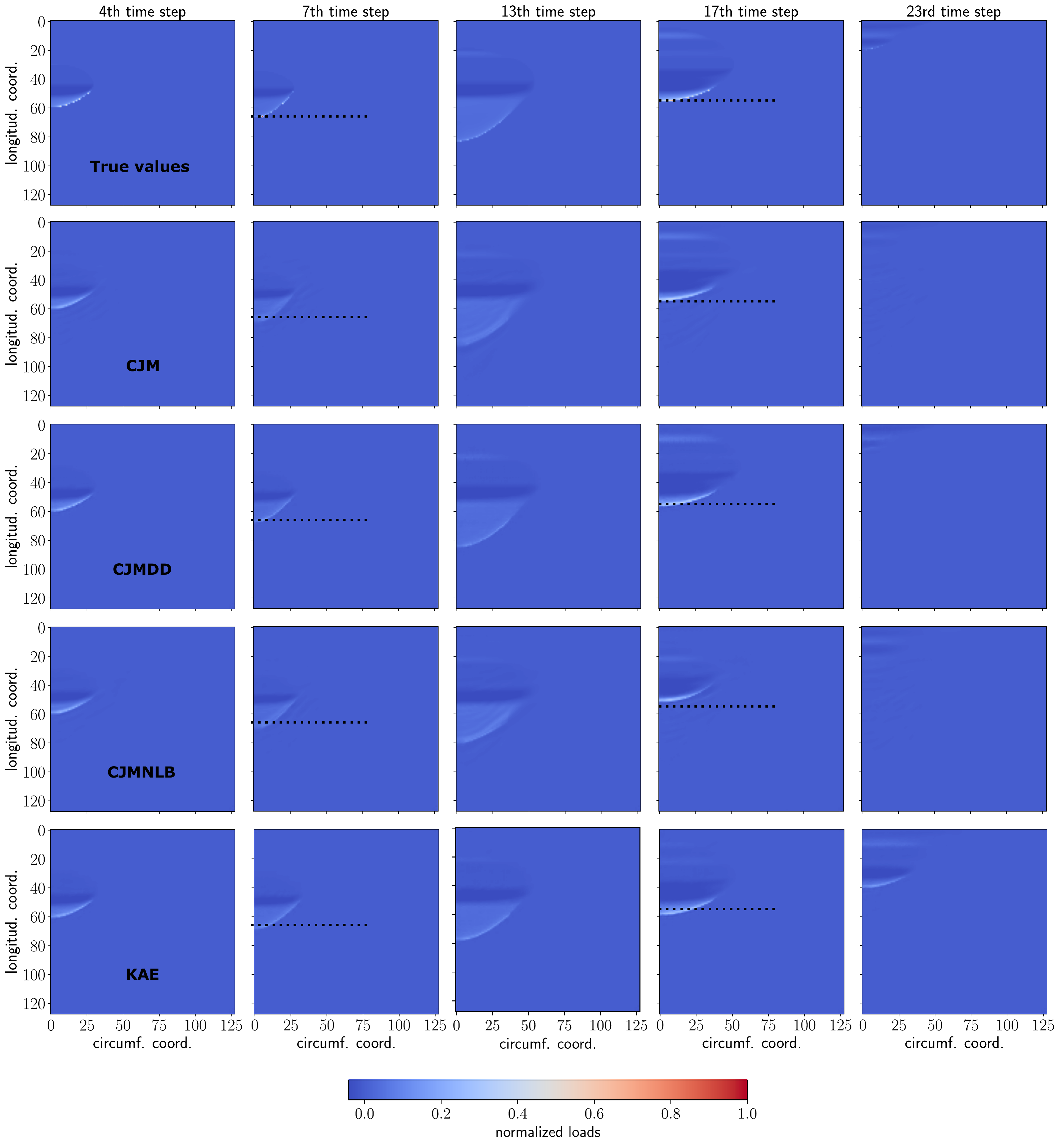}
     \caption{Comparison of unfiltered numerical and surrogate model load predictions for selected time steps (second case). Loads are normalized by the maximum load of $\SI{2370}{kPa}$. The initial $\ell=3$ time steps were used to start the models. Dotted lines cf. Fig \ref{fig:allpredictions1}.}
     \label{fig:unfiltered_allpredictions2}
 \end{figure}

Based on the discussion in Sec. \ref{sec:data_processing},  Figs. \ref{fig:unfiltered_allpredictions1} and \ref{fig:unfiltered_allpredictions2} depict the pressure load predictions obtained for the two central test cases after the CNN depicted in Table \ref{tab:cnn} was applied in a postprocessing step to map the filtered data back to the unfiltered data. Figs. \ref{fig:unfiltered_allpredictions1} and \ref{fig:unfiltered_allpredictions2} correspond to Figs. \ref{fig:allpredictions1} and \ref{fig:allpredictions2} containing the pressure load predictions for the filtered data. For Fig. \ref{fig:unfiltered_allpredictions2}, the second column refers to the 7th time step instead of the 8th as in Fig. \ref{fig:allpredictions2}, since the peak value of $\SI{2370}{kPa}$ occurs at this time step.
The first three convolutional layers in the CNN are activated using \textsf{LeakyReLU} with slope parameter $\alpha=0.01$ and the last convolutional layer is linear. The CNN was trained for 50 epochs using Adam with a minibatch size of 64.
\begin{table}[h!]
\caption{CNN structure. Conv2D($f$, $k$, $s$) denotes a 2D convolutional layer with $f$ filters of kernel size $k\times k$ and stride $s$ in both dimensions. The shapes do not include the minibatch size. The network is trained on grayscale patches of shape $128\times128$, we use channels last, thus the input is of shape $128\times128\times1$.}
\centering
\begin{tabular}{cccc}
    \hline
    \hline
    Layer & Output shape & Layer & Output shape\\
    \hline
     Input & (128,128,1) &  Conv2D(64, 3, 1) & (128,128,64) \\
     Conv2D(16, 3, 1) & (128,128,16) & Conv2D(1, 3, 1) & (128,128,1) \\
     Conv2D(32, 3, 1) & (128,128,32) & Reshape(128,128) & (128,128) \\
     \hline
     \hline
\end{tabular}
\label{tab:cnn}
\end{table}

\begin{figure}
    \centering
    \includegraphics[width=16cm]{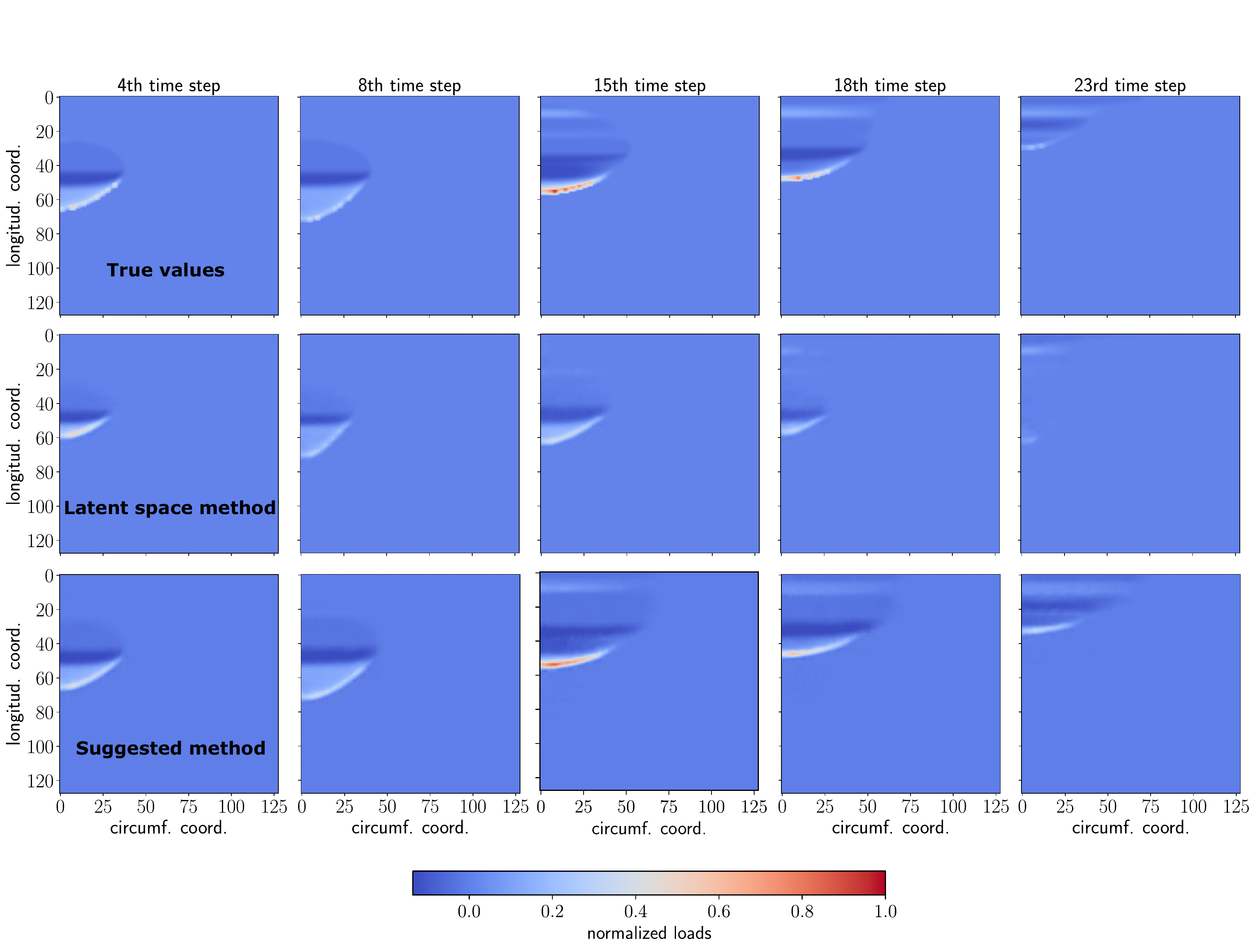}
    \caption{Comparison of two different KAE approaches applied to the first test case.  Results are normalized by the maximum load of 740 kPa. The initial $\ell=3$ time steps of the impact were used to start the models.}
    \label{fig:kae_latent_predictions1}
\end{figure}

\begin{figure}
    \centering
    \includegraphics[width=16cm]{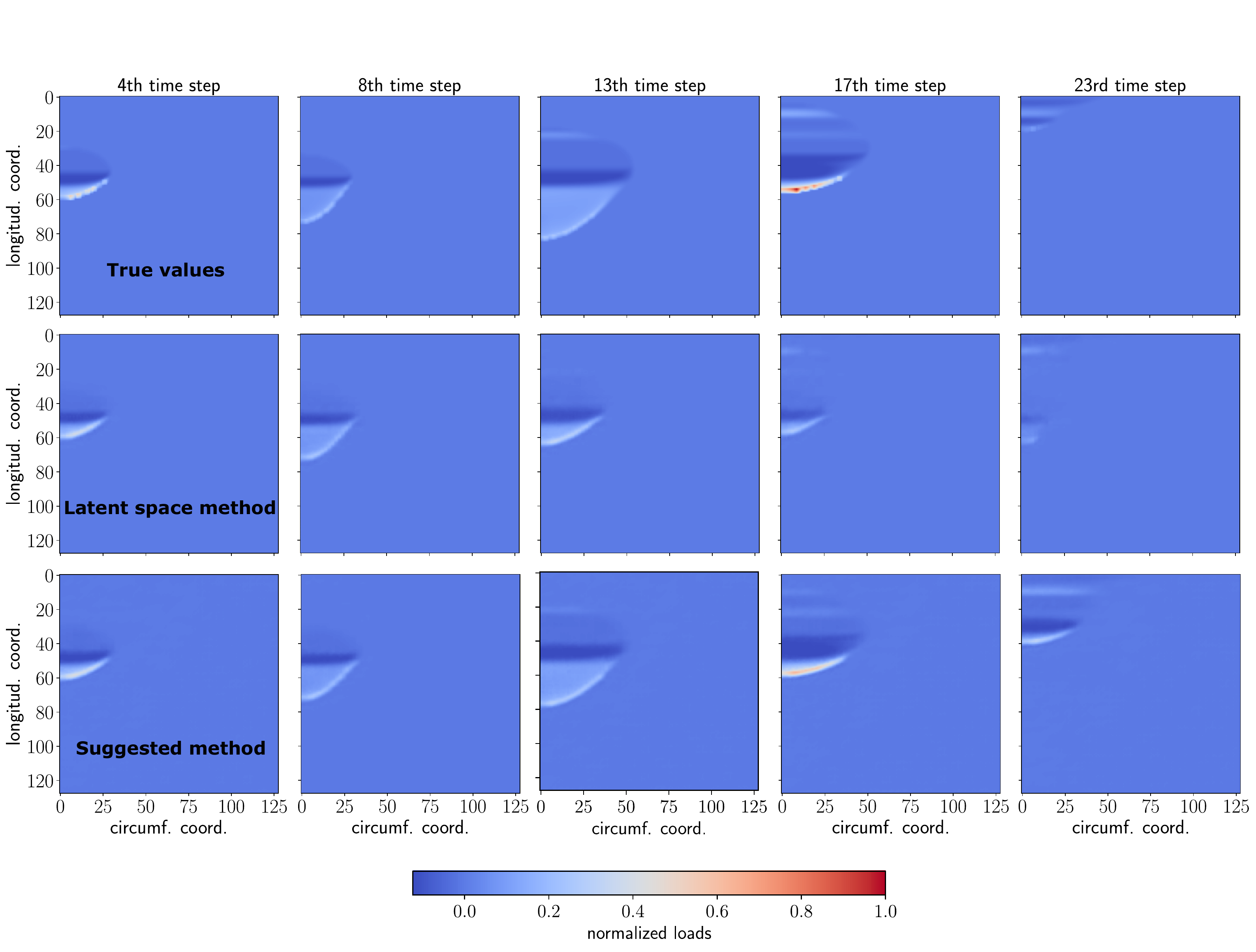}
    \caption{Comparison of two different KAE approaches applied to the second test case. Results are normalized by the maximum load of 810 kPa. The initial $\ell=3$ time steps of the impact were used to start the models.}
    \label{fig:kae_latent_predictions2}
\end{figure}

Figs. \ref{fig:kae_latent_predictions1} and \ref{fig:kae_latent_predictions2} reuse the two focal test cases discussed in detail to demonstrate that mapping out of the latent space each time step leads to better results for the KAE than predicting solely in the latent space, cf. discussion in Sec. \ref{subsec:load_prediction}. Displayed results refer to the best results of the five trained KAE versions. The predictions in the second rows of the figures were performed solely in the latent space and afterwards decoded. The predictions in the third rows were obtained using the full model each time step.

\begin{figure}
    \centering
    \includegraphics[width=14cm]{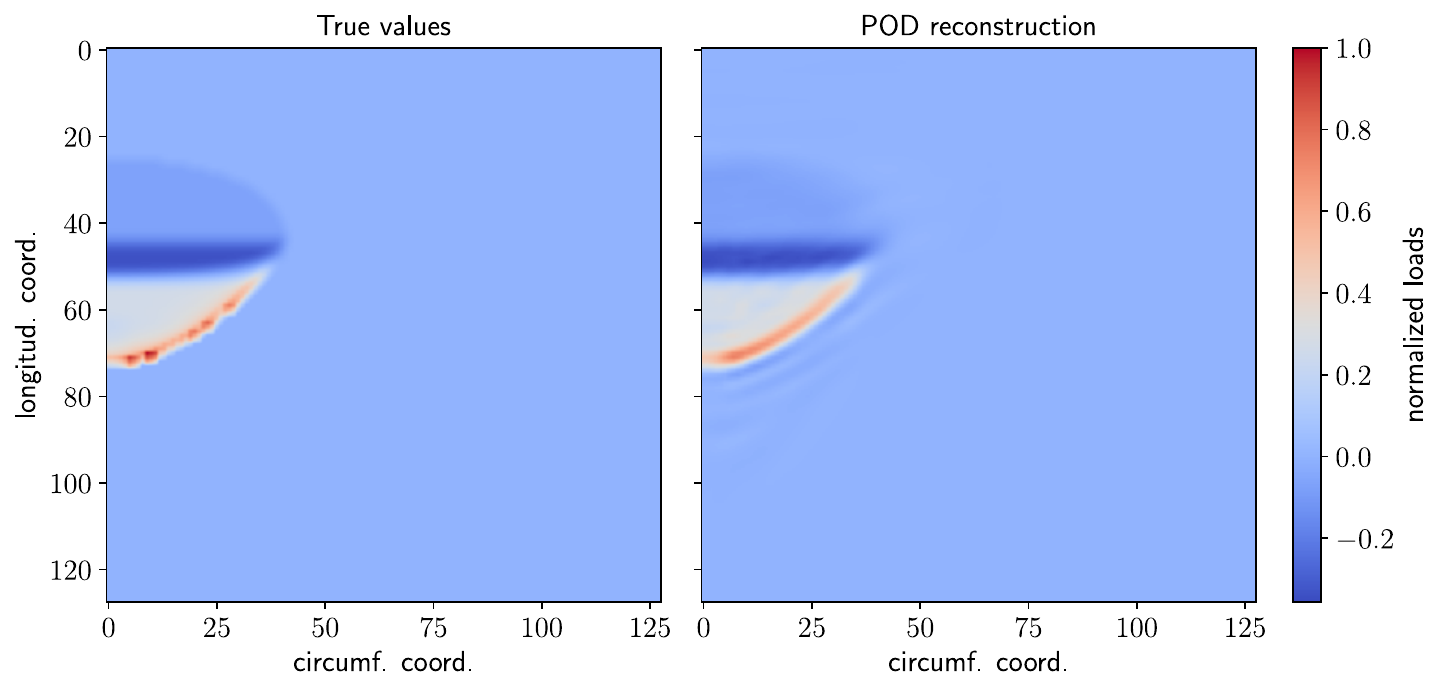}
    \caption{Comparison of numerical load prediction
and POD reconstruction using 100 coefficients for the 8th time step of the first test case. Loads are normalized by the maximum load of $\SI{298}{kPa}$.}
    \label{fig:pod_reconstruction}
\end{figure}

Fig. \ref{fig:pod_reconstruction} illustrates oscillations in the load reconstruction of a POD with 100 modes, cf. discussion in Sec. \ref{subsec:load_prediction}.

\section*{Acknowledgements}
H.S., M.Ü., \ and T.R.\ acknowledge support by the German Federal Ministry for Economic Affairs and Energy under aegis of the ''Luftfahrtforschungsprogramm LuFo VI'' projects  INSIDE
(grant 20Q1951B) and HYMNE (grant 20E2218A) 
This paper is  a contribution to the research training group 
RTG 2583 on ''Modeling, Simulation and Optimization of Fluid Dynamic Applications''  funded by the Deutsche Forschungsgemeinschaft (DFG). The authors acknowledge the support of TU Hamburg's Machine Learning in Engineering (MLE) initiative. The authors would like to thank the anonymous referees for their insightful comments and suggestions that improved the presentation of the results.

\bibliography{./references.bib}

\end{document}